\documentclass[10pt,journal,compsoc]{IEEEtran}
\ifCLASSOPTIONcompsoc
  \usepackage[nocompress]{cite}
\else
  \usepackage{cite}
\fi
\ifCLASSINFOpdf
\else
\fi
\usepackage{graphicx}
\usepackage{amsmath}
\usepackage{amssymb}
\usepackage{booktabs}
\usepackage{multirow}
\usepackage{mathrsfs}
\usepackage{amsfonts}
\usepackage{mathabx}
\usepackage{epsfig}
\usepackage{graphicx}
\usepackage{amsmath}
\usepackage{amssymb}
\usepackage{enumitem}
\usepackage{comment}
\usepackage{diagbox}
\usepackage{multirow}
\usepackage{subcaption}
\usepackage{array}
\usepackage{placeins}
\usepackage{bm}
\usepackage{color}
\usepackage{url}
\usepackage{color}
\newcommand{\red}[1]{\textcolor{red}{#1}}

\usepackage{algorithm}  
\usepackage[noend]{algpseudocode} 
\usepackage{amsmath,amsfonts}

\usepackage{algorithmicx,algorithm}

\newcommand{\myblue}[1]{{\color{black}{#1}}}

\usepackage{amsthm}

\def \R {\mathbb{R}}
\def \S {\mathcal{S}}

\newtheorem{thm}{Theorem}

\usepackage{hyperref}

\definecolor{lzycolor}{rgb}{0.1,0.2,0.7}

\definecolor{zqlcolor}{rgb}{0.0,0.0,0.0}

\begin{document}
%\title{Practical Model Repairing Against Both Backdoor and Adversarial Attacks}
\title{Towards Unified Robustness Against Both Backdoor and Adversarial Attacks}

\author{Zhenxing Niu$^*$,~\IEEEmembership{Member,~IEEE},
        Yuyao Sun$^*$,~\IEEEmembership{Student Member,~IEEE},
        Qiguang Miao,~\IEEEmembership{Member,~IEEE},\\
        Rong Jin,~\IEEEmembership{Fellow,~IEEE},
        Gang Hua,~\IEEEmembership{Fellow,~IEEE}%
\IEEEcompsocitemizethanks{
\IEEEcompsocthanksitem Z. Niu, Y. Sun and Q. Miao are with the Key Laboratory of Smart HumanComputer Interaction and Wearable Technology of Shaanxi Province, Xidian University, China.
E-mail: \{zxniu, yysun, qgmiao\}@xidian.edu.cn. (\thefootnote{*} Equal contribution. Corresponding author: Zhenxing Niu.)\protect
\IEEEcompsocthanksitem R. Jin is with DAMO Academy, Alibaba Group, Beijing, China.
E-mail: rongjin@cse.msu.edu.\protect
\IEEEcompsocthanksitem G. Hua is with Wormpex AI Research, Bellevue, WA 98004, USA.
E-mail: ganghua@gmail.com.\protect
}% <-this % stops an unwanted space
}

\markboth{IEEE Transactions on Pattern Analysis and Machine Intelligence,~Vol.~XX, No.~XX, XXXX~XXXX}{Shell \MakeLowercase{\textit{et al.}}: Weakly Supervised Temporal Action Localization through Contrast based Evaluation Networks}



\IEEEtitleabstractindextext{%
\begin{abstract}
Deep Neural Networks (DNNs) are known to be vulnerable to both backdoor and adversarial attacks. In the literature, these two types of attacks are commonly treated as distinct robustness problems and solved separately, since they belong to training-time and inference-time attacks respectively. However, this paper revealed that there is an intriguing connection between them: (1) planting a backdoor into a model will significantly affect the model's adversarial examples; (2) for an infected model, its adversarial examples have \emph{similar features} as the triggered images. Based on these observations, a novel \textbf{P}rogressive \textbf{U}nified \textbf{D}efense (PUD) algorithm is proposed to defend against backdoor and adversarial attacks \emph{simultaneously}. Specifically, our PUD has a progressive model purification scheme to jointly erase backdoors and enhance the model's adversarial robustness. At the early stage, the adversarial examples of infected models are utilized to erase backdoors. With the backdoor gradually erased, our model purification can naturally turn into a stage to boost the model's robustness against adversarial attacks. Besides, our PUD algorithm can effectively identify poisoned images, which allows the initial extra dataset not to be \emph{completely} clean. Extensive experimental results show that, our discovered connection between backdoor and adversarial attacks is ubiquitous, no matter what type of backdoor attack. The proposed PUD outperforms the state-of-the-art backdoor defense, including the model repairing-based and data filtering-based methods. Besides, it also has the ability to compete with the most advanced adversarial defense methods. The code is available \href{https://github.com/John-niu-07/PUD}{here}.
\end{abstract}

}

\maketitle

\IEEEdisplaynontitleabstractindextext

% \fi
\IEEEpeerreviewmaketitle

\IEEEraisesectionheading{\section{Introduction}\label{sec:introduction}}


\IEEEPARstart{D}eep neural networks (DNNs) have been widely adopted in many safety-critical applications (\emph{e.g.}, face recognition and autonomous driving), thus more attention has been paid to the security of deep learning. It has been demonstrated that DNNs are prone to some potential threats both at inference time and at training time. At inference time, it is found that we can easily fool a well-trained model into making incorrect predictions with small adversarial perturbations, which is called an adversarial attack~\cite{goodfellow2014explaining,szegedy2014intriguing}. 

On the other hand, \myblue{it is} found that DNNs are also vulnerable to another type of attack, called a backdoor attack~\cite{li2020backdoorsurvey}, which happens at model training time. Specifically, a backdoor attack attempts to plant a backdoor into a model in the training phase, so that the backdoored/infected model would misclassify any test image as the \emph{ backdoor target-label} whenever a predefined \emph{trigger} (\emph{e.g.}, several pixels) is embedded in the images (called \emph{triggered/poisoned images}). 

\begin{figure}[t]
\vspace{-3.5em}
\centering
\subfloat[A histogram of predicted labels of adversarial examples w.r.t. \textbf{infected} model]{
\label{cifar_hist:1}
\includegraphics[width=0.45\linewidth]{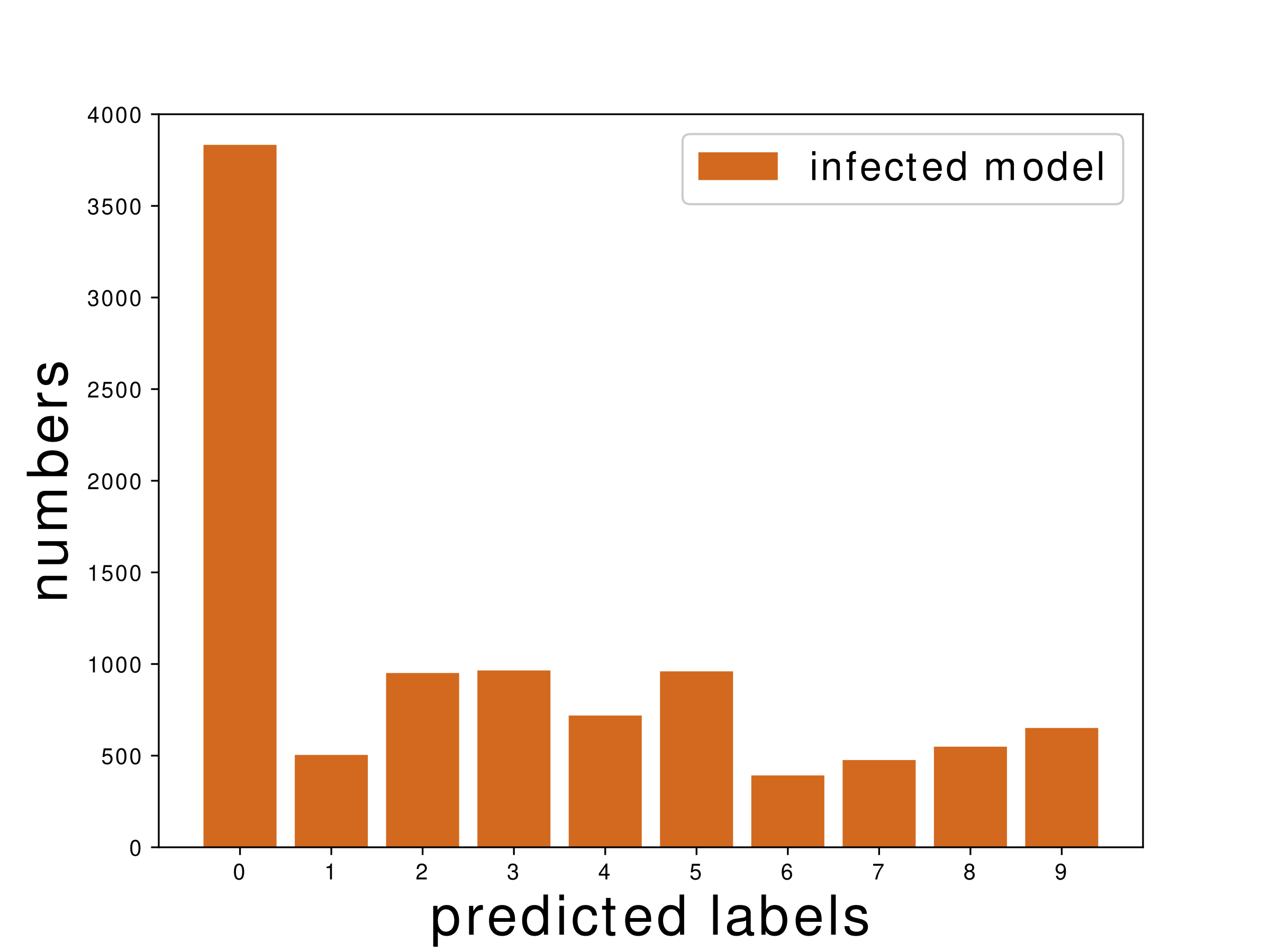}}
\quad
\subfloat[A histogram of predicted labels of adversarial examples w.r.t. \textbf{benign} model]{
\label{cifar_hist:2}
\includegraphics[width=0.45\linewidth]{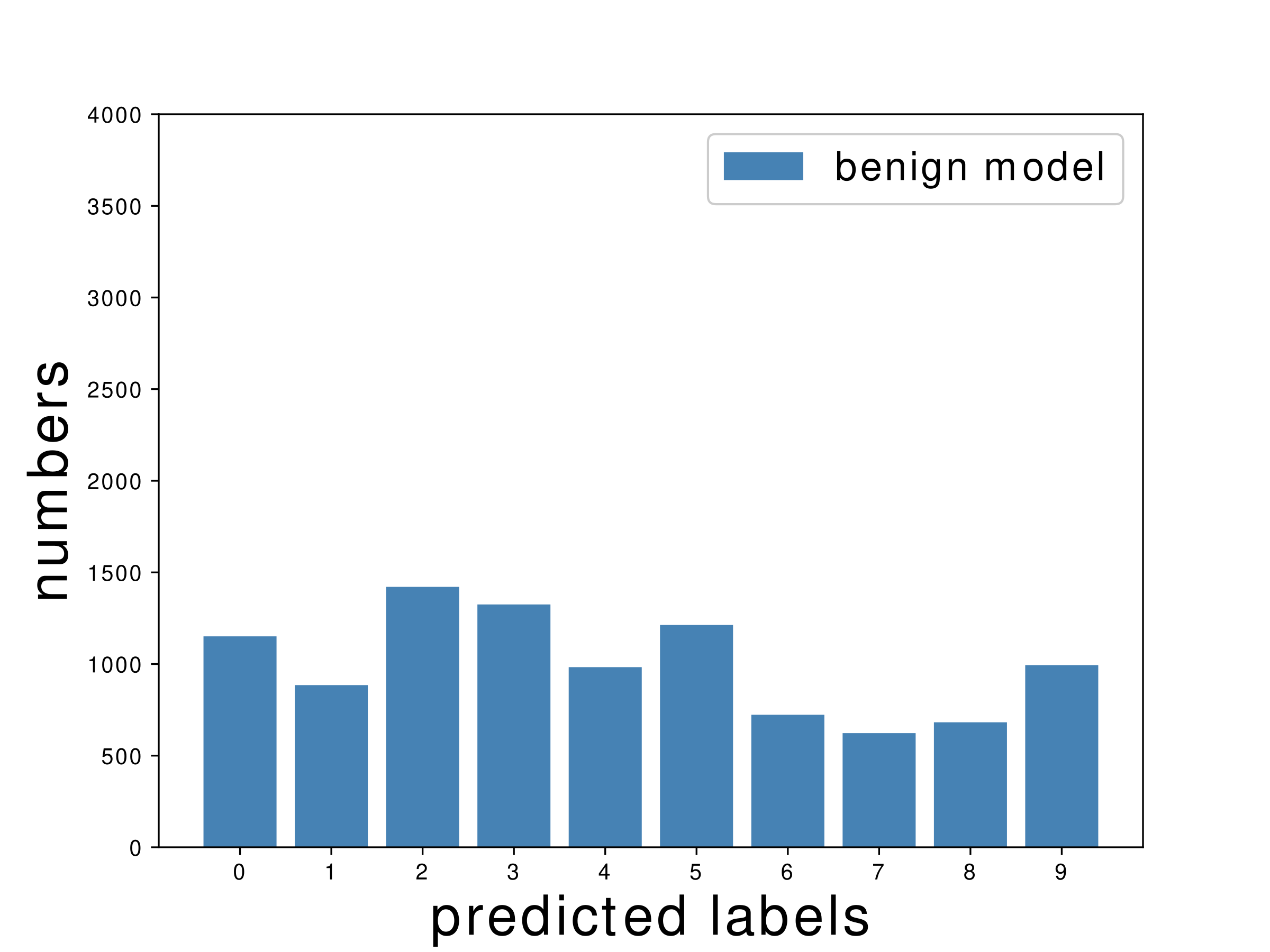}}
\caption{With $10,000$ images randomly sampled from CIFAR-10:  
(a) For an \textbf{infected} model (whose backdoor target-label was predetermined as \emph{`class 0'} in this case), its adversarial examples are highly likely classified as the target-label (\emph{i.e.,} \emph{`class 0'}). (b) But for a \textbf{benign} model, its adversarial examples could be classified as any
class with almost equal probability.}\label{cifar_hist}
\vspace{-1.5em}
\end{figure}

Due to the inherent differences between adversarial and backdoor attacks, they are often treated as two separate problems and solved separately in the literature. As reviewed in the related work section, there is a great deal of work on adversarial attacks, backdoor attacks and defenses against them separately. However, the connection between them has been less studied. Furthermore, there are fewer methods that are able to defend against both adversarial and backdoor attacks simultaneously.

In this paper, we reveal that
there is an underlying connection between adversarial and backdoor attacks. We discover this connection by performing an untargeted adversarial attack on an infected model. Note that in all previous work adversarial attacks were performed on clean/benign models instead of infected models. Specifically, after generating adversarial examples for an \textbf{infected model} (Fig.\ref{cifar_hist:1}), we surprisingly observe that these adversarial examples are highly likely to be classified as the \emph{backdoor target-label}  (which was predetermined when planting backdoors at the training stage). In contrast, for a \textbf{benign model} (Fig.\ref{cifar_hist:2}), its adversarial examples can be classified as \emph{any class} with almost equal probability, \emph{i.e.}, obeying a uniform distribution. 

Following this cue, we reveal that: (1) After planting a backdoor in a model, the features of the adversarial images change significantly. As shown in Fig.\ref{backdoor_feat}, the features of the adversarial image $\widetilde{\bm{x}}'$ are significantly different from the features of the original adversarial image $\widetilde{\bm{x}}$ (which is generated from a clean model). (2) Surprisingly, \textbf{the features of $\widetilde{\bm{x}}'$ are very similar to the features of the triggered image $\bm{x^t}$}. It indicates that both the $\widetilde{\bm{x}}'$ and $\bm{x^t}$ tend to activate the \emph{same} DNN neurons. Revealing such an intriguing connection gives us the opportunity to defend against both backdoor and adversarial attacks simultaneously.

In recent years, backdoor attacks have made great advances, evolving from visible trigger~\cite{gu2017badnets} to invisible trigger~\cite{chen2017blend,liu2020reflection,nguyen2021wanet}, and from poisoning-label to clean-label attacks~\cite{barni2019sig}. For example, WaNet~\cite{nguyen2021wanet} uses an affine transformation as a trigger embedding mechanism, which significantly improves the invisibility of the trigger. In contrast, the backdoor defense research has lagged behind. 
%The backdoor defenses can be roughly categorized as model repairing and data filtering methods. Model repairing methods aim to erase backdoors from a given infected model, where the original poisoned training data is usually unavailable. Note that even for the state-of-the-art model repairing methods~\cite{liu2017finetuning,liu2018finepurning,li2020nad}, most of them can be evaded by advanced modern backdoor attacks. Besides, an extra \emph{clean dataset} is often required by these methods, which make them unpractical to realistic backdoor defense scenarios in real-world applications. 
Even for the state-of-the-art backdoor defense~\cite{liu2017finetuning,liu2018finepurning,li2020nad}, most of them can be evaded by the advanced new backdoor attacks~\cite{nguyen2020inputaware,nguyen2021wanet}. Moreover, an extra \emph{clean} dataset is often required by those defense methods.  

On the other hand, many methods have been proposed to defend against adversarial attacks. For example, adversarial training~\cite{madry2018towards} is the most representative one. However, if the original training data is poisoned, adversarial training cannot \myblue{defend against} backdoor attacks~\cite{Trade-off, Manoj}. 

In this paper, based on our discovered connections between backdoor and adversarial attacks, we propose a unified defense method to defend against backdoor and adversarial attacks \emph{\textbf{simultaneously}}, called \textbf{Progressive Unified Defense (PUD)}. Our PUD has a progressive model purification scheme to jointly erase backdoors and enhance the model's adversarial robustness. At the early stage, the adversarial examples of infected models are utilized to erase backdoors. With the backdoor gradually erased, our model purification can naturally turn into a stage to focus on boosting the model's adversarial robustness. 

In particular, \myblue{when} erasing backdoors, our approach does not require the extra dataset to be \emph{completely} clean. To this end, our approach has a `model-data' alternative purification procedure: we first use the poisoned extra dataset to purify the infected model, \emph{i.e.}, erasing backdoors (dubbed `model purification'). After that, the purified model is further used to purify the extra dataset, \emph{i.e.}, identifying and removing poisoned images (dubbed `data purification'). With a few alternative iterations, both the infected model and the poisoned extra dataset can be gradually purified.

%Our approach could not only defend against modern backdoor attacks but also can work \emph{without} a \emph{clean} extra dataset. Specifically, we just assume to be able to access \emph{a small extra dataset which may contain poisoned images}. It can be a small subset of training data in practice. At the beginning, it is regarded as the initial extra dataset. In the following iterations, we first use the extra dataset to generate adversarial examples, and then use the generated adversarial examples to purify the infected model. After that, the purified model is further used to purify the extra dataset (identify and remove poisoned images). With such an iterative procedure, both the infected model and the extra dataset are progressively purified, as shown in Fig.\ref{mPBE}. 

Specifically, regarding the model purification, we generate adversarial examples and use them to fine-tune the infected model. Why can we purify the infected model via \textbf{fine-tuning with adversarial examples}? It is because these adversarial images, on the one hand, come from arbitrary classes (annotated with different class labels), and on the other hand, have similar features as the triggered images (\emph{i.e.,} our observations). As a result, the fine-tuning with adversarial examples works like fine-tuning with the \emph{`triggered images' annotated with different class labels}. It breaks the foundation of backdoor attacks that aim to establish a strong correlation between triggered images and the backdoor target-label~\cite{li2020nad}.

\myblue{With the backdoor being gradually erased}, the infected model gradually becomes a clean model, and our adversarial fine-tuning step (\emph{i.e.,} Step-1 in Algorithm 1) naturally becomes a normal adversarial training procedure, which benefits the model robustness against adversarial attacks. Therefore, the main advantage of our approach is that \emph{it can defend against both backdoor and adversarial attacks}. Moreover, to further enhance the model's adversarial robustness, a teacher-student mechanism~\cite{Meanteachers} is adopted by our PUD. It can not only boost the backdoor erasing, but also significantly improve the model's adversarial robustness. %Besides, we find that the model robustness suffers from the \emph{robust overfitting} issue~\cite{Overfitting}. Thus, we adopt an early stopping strategy to mitigate such limitations.

%But our empirical experiments tell us that even the state-of-the-art data filtering methods \emph{SPECTRE} cannot effectively identify poisoned images, especially when the backdoor attack is strong (\emph{i.e.,} with a large poisoning ratio). In contrast, our strategy of measuring prediction similarity could better handle such situations. More experiments turn out there is complementary between our approach and SPECTRE. Thus, we combine them together to identify clean images in our mPBE method. Note that in our original PBE method, we just use our strategy of measuring prediction similarity. There is a significant improvement from PBE to mPBE due to the combination of such two strategies. 

As we know, most existing backdoor erasing methods assume that the extra dataset is clean. However, we allow it to contain \emph{\textbf{poisoned}} images, \emph{i.e.,} our PUD can erase backdoors even if the extra dataset is not \emph{completely} clean. In practice, it is easier to get a poisoned extra dataset than a clean extra dataset, since it is often unknown to a defender which images are poisoned. The ability to erase backdoors without clean data is because our approach can purify the extra dataset by itself. Specifically, we propose a prediction consistency-based strategy to identify poisoned images in the extra dataset. Furthermore, we combine it with another data filtering strategy \emph{SPECTRE}. The two strategies complement each other and significantly improve the effectiveness of data purification. To the best of our knowledge, our approach is the first work to defend against backdoor attacks without an extra clean dataset. 

Furthermore, the identified poisoned images in the extra dataset are not just discarded. Instead, we utilize them to further remove the backdoor through a \emph{machine unlearning} scheme~\cite{ABL,Unlearning,MUnlearning}, which can significantly reduce ASR. Thus, both clean and poisoned images in the extra dataset can be fully utilized in our approach.

%-------------------------------------------------------------------------

%contributions
In summary, the main contributions of this paper include (1) revealing an underlying connection between backdoor and adversarial attacks (\emph{i.e.,} for an infected model, its adversarial examples have similar features as the triggered images), and providing a theoretical analysis to justify this observation; (2) proposing a unified defense algorithm to improve the model robustness against both backdoor and adversarial attacks; (3) 
be able to defend against the state-of-the-art backdoor attacks, even when an extra dataset is not completely clean.

% 联合训练变成迭代训练以克服初始化问题
% 提供了更多的细节
% 通过实验，公平比较了阈值方法和本方法的区别，并且提供了更多的可视化结果
% 添加了新的实验，包括在新的feature下的实验，对超参数的实验

This paper extends our work, \myblue{Progress Backdoor Erasing (PBE)} \cite{PBE} in three ways. First, a teacher-student mechanism is proposed to improve the model robustness against both backdoor and adversarial attacks. Second, a data filtering scheme is adopted to complement the data purification of PBE. Third, different from PBE which only uses clean images to erase backdoors, our PUD also utilizes the identified \emph{poisoned} images to further erase backdoors through \emph{machine unlearning} schemes. 

Due to the three improvements, (1) compared to PBE, our PUD can not only defend against backdoor attackes, but also defend against adversarial attacks; (2) PBE needs to access the training data, but our PUD does not. The data purification of PBE is weak, so it needs to use the training data to augment/update its extra dataset. However, PUD has a strong data purification mechanism, so the given small extra dataset is sufficient for defense. This makes our PUD more practical for realistic scenarios.

%our PUD only needs \emph{a small} extra dataset while the PBE requires to access \emph{all} poisoned training data. 

%The rest of the paper is organized as follows. After reviewing related work in Section~\ref{sec:relatedwork}, the connection between adversarial and backdoor attack are presented in Section~\ref{sec:obs}. And then, we introduce how to leverage the discovered connection to design a unified defense algorithm against both backdoor and adversarial attacks in Section~\ref{sec:alg}.

%------------------------------------------------------------------------
%------------------------------------------------------------------------

\section{Related Work}\label{sec:relatedwork}
\subsection{Backdoor Attack}

In general, backdoor attacks aim to plant backdoors in DNNs during the training process, so that the infected DNNs behave normally on benign images, but their predictions for \emph{triggered/poisoned images} are maliciously changed to a predetermined class label, \emph{i.e.,} backdoor target-label. Currently, poisoning training images is the easiest and most widely adopted method for planting backdoors, which is achieved by editing images with a specific trigger. Therefore, the trigger embedding mechanism is the key to backdoor attacks. Stealth is the most desired property for embedding triggers, which has motivated the evolution from poisoning label to clean-label attacks~\cite{barni2019sig} (only the images from the backdoor target-label class are triggered) and the evolution from visible triggers~\cite{gu2017badnets} to invisible triggers~\cite{chen2017blend,liu2020reflection,nguyen2021wanet}.

Instead of using a predetermined trigger (such as a patch (Badnet)~\cite{gu2017badnets}, a sinusoidal strip (SIG)~\cite{barni2019sig}, or a blending image (Blend)~\cite{chen2017blend}), learning a dynamic trigger pattern can significantly improve the stealth of backdoor attacks. This has evolved from learning an uniform trigger pattern to learning different trigger patterns for different images. 
For example, the former emphasize joint optimization of trigger patterns and backdoored models, such as TrojanNN~\cite{Trojaning} and EvilTwins~\cite{EvilTwins}. 
The latter move forward to make trigger patterns adaptive to image content, which aims to generate image-specific trigger patterns, such as input-aware dynamic patterns (DyAtt)~\cite{nguyen2020inputaware}, natural reflection~\cite{liu2020reflection} and image warping (WaNet)~\cite{nguyen2021wanet}. \myblue{In addition, there is another type of attack known as a code poisoning attack (\emph{i.e.}, Blind Attack)~\cite{bagdasaryan2021blind}, which aims to implant a backdoor through malicious code injection rather than poisoning training data.
AdvDoor~\cite{zhang2021advdoor} employs the targeted universal adversarial perturbation to generate a trigger, enabling it to evade trigger detection methods like Spectral. \cite{tran2018spectral} proposes a handcrafted attack that directly manipulates a model’s weights.} A survey of backdoor attacks can be found in~\cite{li2020backdoorsurvey}.

%-------------------------------------------------------------------------
\subsection{Backdoor Defense}
Backdoor defense can be broadly divided into four categories: (1) \emph{backdoor detection} methods~\cite{wang2019neuralcleanse,kolouri2020universal}, which aim to detect the presence of a backdoor in a suspect model. Although they perform quite well in distinguishing whether a model has been backdoored, the backdoor still remains in the infected model. (2) \emph{data filtering} methods~\cite{chen2019detectingcluster,peri2020deepknn,tran2018spectral, hayase2021spectre} that attempt to detect and remove or sanitize
malicious/poisoned training data; (3) \emph{robust training} methods~\cite{Trade-off, Manoj, Ezekiel, Kill, Effectiveness}, which attempt to design a training routine that yields robust models even on malicious training data; and (4) \emph{model repairing} methods~\cite{liu2017finetuning,liu2018finepurning,li2020nad,ANP}, which assume to just have an infected model, and need to repair the infected model. Note that both data filtering and robust training methods perform defense \emph{during} training, thus they assume that the model training procedure is controlled by the defender. In contrast, model repair methods perform defense \emph{after} training, without the requirement of controlling the training procedure. This makes it more suitable for practical applications, and it becomes the mainstream of modern backdoor defense. For example, it can defend against input-aware backdoor attacks~\cite{nguyen2020inputaware, nguyen2021wanet} where the training procedure is controlled by the adversary instead of the defender.

Regarding the model repairing methods, a straightforward method is to directly fine-tune the infected model with an extra clean dataset\cite{liu2017finetuning}. After that, Fine-Pruning \cite{liu2018finepurning} is proposed to use neural pruning to remove backdoor neurons. In~\cite{li2020nad}, Neural Attention Distillation (NAD) is proposed to erase backdoor by using knowledge distillation. Later, Adversarial Neuron Pruning (ANP)~\cite{ANP} is proposed to prune backdoor neurons by perturbing model weights. \myblue{In~\cite{zeng2021adversarial}, the Implicit Backdoor Adversarial Unlearning (I-BAU) is proposed to achieve a good balance between model accuracy and robustness.}
To further improve the efficiency of backdoor erasing, some 
%trigger synthesis-based 
\myblue{reverse engineering-based} methods are proposed. Neural Cleanse (NC)~\cite{wang2019neuralcleanse} and Artificial Brain Stimulation (ABS)~\cite{liu2019abs} are proposed to recover the backdoor trigger first, and use the recovered trigger to erase the backdoor. 
\myblue{Recently, ~\cite{sun2023single} introduces a method capable of recovering the trigger with only one clean image. It leverages the robust smoothed version of the backdoored classifier, implying a connection between adversarial robustness and backdoor attacks.}
However, these methods have two shortcomings: (1) all of them require an extra clean dataset to perform the defense; (2) most of them are only able to handle a fixed (even learnable) trigger since they need to explicitly recover it. As a result, they cannot effectively defend against dynamic/content-aware attacks such as DyAtt~\cite{nguyen2020inputaware}, WaNet~\cite{nguyen2021wanet}.
In contrast, our approach does not require an extra clean dataset; on the other hand, our approach does not need to recover trigger patterns, so it can deal with dynamic attacks.

%-------------------------------------------------------------------------
\subsection{Adversarial Attack and Defense}
%Adversarial attack~\cite{goodfellow2014explaining,szegedy2014intriguing,ilyas2019adversarial} is a type of inference-time attack that aims to fool a trained model into making incorrect predictions (\emph{i.e.,} untargeted adversarial attack) or predicting the input as a particular label (\emph{i.e.,} targeted adversarial attack). The first proposed adversarial attack is the Fast Gradient Sign Method (FGSM)~\cite{goodfellow2014explaining} which generates adversarial examples with a single normalized gradient step. This was followed by the Basic Iterative Method (BIM)~\cite{BIM}, which takes several smaller gradient steps. Finally, they are all unified and formulated as the Projected Gradient Descent (PGD)~\cite{madry2018towards} attack by referring to the optimization procedure used to search for norm-bounded perturbations. 

%On the other hand, many defenses against adversarial attacks are also proposed. Adversarial Training (AT)\cite{madry2018towards} achieves defense by feeding adversarially perturbed examples back into the training data, which is considered to be one of the most effective defense methods. 

\myblue{Since Szegedy et al. observed that neural networks are highly vulnerable to adversarial examples, the art of generating adversarial examples has received a lot of attention. The Fast Gradient Sign Method (FGSM)~\cite{goodfellow2014explaining} was first proposed to find the image perturbation with a single normalized gradient step. The adversary was later extended to take multiple smaller steps, known as the Basic Iterative Method (BIM)~\cite{BIM}. Finally, they are all unified and formulated as Projected Gradient Descent (PGD)~\cite{madry2018towards} attack by referring to the optimization procedure used to search for norm-bounded perturbations. Instead of optimization under a given perturbation constraint, C\&W attack~\cite{carlini2017evaluating} aims to find the smallest successful adversarial perturbation. Recently, Auto-PGD~\cite{croce2020reliable} is proposed to optimize attack strength by adapting the step size across iterations depending on the overall attack budget and progress of the optimisations.}

\myblue{On the other hand, several methods have been proposed to defend against such attacks. Adversarial Training (AT)\cite{madry2018towards} achieves defense by incorporating adversarially perturbed examples back into the training data, and it is considered one of the most effective defense methods. 
Later, the AT method has been augmented in various ways, including changes in the attack procedure (\emph{e.g.}, by incorporating momentum~\cite{momentum}), loss function (\emph{e.g.}, logit pairing~\cite{pairing}), or model architecture (\emph{e.g.}, feature denoising~\cite{denoising}). Lately, another notable work is TRADES~\cite{TRADES}, which aims to balance the trade-off between standard and robust accuracy. MART~\cite{Wang2020Improving} is proposed to address this trade-off by using boosted loss functions. Recently, \cite{gowal2021uncovering} performed a systematic analysis of many different aspects surrounding adversarial training that can affect the robustness of trained networks.}

%-------------------------------------------------------------------------
\subsection{Connections between Adversarial and Backdoor Attacks}
There have been some attempts to consider adversarial and backdoor attacks together, and to use them to help each other. 
Recently, \cite{10.1145/3372297.3417231} proposed using a `trapdoor' to detect adversarial examples. It showed that a particular trapdoor could lead to producing adversarial examples similar to trapdoors in the feature space. However, this differs from our work in that it aims to detect adversarial examples while our approach aims to erase backdoors. 

In addition, some methods~\cite{Trade-off, Manoj, Ezekiel, Kill, Effectiveness} attempt to adopt the adversarial training as a backdoor robust training method. However, they pay no attention to the connection between two types of attacks. To the best of our knowledge, there has been less work focused on finding the underlying connections between them. 
%-------------------------------------------------------------------------

\section{The Connection between Backdoor and Adversarial Attacks}\label{sec:obs}
In this section, we will describe how we discover the underlying connection between backdoor and adversarial attacks step by step. First, we provide some background on backdoor and adversarial attacks. Second, we make an interesting observation about adversarial examples and triggered images. Third, we reveal the 
connection between backdoor and adversarial attacks. Finally, we provide a theoretical analysis to explain such connections. 

\subsection{Preliminaries}
\subsubsection{Backdoor Attack}
We focus on backdoor attacks on image classification. Let $D_{\mathrm{train}} =\{(\bm{x}_i; y_i)\}_{i = 1}^N$ be the clean training data and $f(\bm{x};\theta)$ be the benign CNN model with parameter $\theta$. %There are $K$ image classes and $y_i \in \{1,\dots,K\}$. 

For a backdoor attack, we define or learn a trigger embedding function $\bm{x}^t = \mathrm{Trigger}(\bm{x})$ that can convert a clean sample $\bm{x}_i$ into a triggered/poisoned sample $\bm{x}^t_i$. Given a target-label $l$, we can poison a small part of the training samples, \emph{i.e.,} replace $(\bm{x}_i,y_i)$ with $(\bm{x}^t_i,l)$, which produces poisoned training data $D'_{\mathrm{train}}$. Training with $D'_{\mathrm{train}}$ results in the infected model $f(\bm{x};\theta')$. Note that different attacks will define different trigger embedding functions $\mathrm{Trigger}(\cdot)$. 

At test time, if a clean input $(\bm{x},y) \in D_{\mathrm{test}}$ is fed to the infected model, it should be correctly predicted as $y$. However, for a triggered sample $\bm{x}^t$, its prediction changes to  the target-label $l$. In particular, backdoor attacks can be divided into two categories according to the choice of target-labels: (1)All-to-one attack: the target-labels for all samples are set to $l$; (2)All-to-all attack: the target-labels for different classes could be set differently, such as $y+1$, \emph{i.e.},
\begin{align}
&\text{All-to-one attack: } 
\left\{
\begin{array}{l}
    f(\bm{x};\theta') =y; \\
    f(\bm{x}^t;\theta') =l, \bm{x}^t= \mathrm{Trigger}(\bm{x})
\end{array}
\right.\\
&\text{All-to-all attack: }
\left\{
\begin{array}{l}
    f(\bm{x};\theta') =y; \\
    f(\bm{x}^t;\theta') =y+1, \bm{x}^t= \mathrm{Trigger}(\bm{x})
\end{array} \notag
\right.
\end{align}

\subsubsection{Backdoor Defense}\label{sec:def_setting} 

We assume a typical defense setup where the defender has an infected model $f(\bm{x};\theta')$ as well as an extra clean dataset $D_{\mathrm{ext}}$. The goal of the backdoor defense is to \emph{erase} the backdoor trigger from the model while preserving the performance of the model on clean samples. In other words, we want to obtain a cleaned/purified model $f(\bm{x};\theta^c)$ such that:
\begin{align}
\left\{
\begin{array}{l}
    f(\bm{x};\theta^c) =y; \\
    f(\bm{x}^t;\theta^c) =y, \bm{x}^t= \mathrm{Trigger}(\bm{x})
\end{array}
\right.
\end{align}

\subsubsection{Untargeted Adversarial Attack}\label{sec:uaa} 
Untargeted adversarial attack aims to find the best perturbation $\bm{r}$ so that the adversarial examples $\widetilde{\bm{x}}=\bm{x+r}$ are misclassified, \emph{i.e.,} the loss $L(\widetilde{\bm{x}},y)$ is maximized with respect to $\bm{r}$, as follows:
\begin{align}
    &\mathop{\max}_{\bm{r}} L(\widetilde{\bm{x}},y;\theta) \label{eq:adv} \\
    &\text{s.t. }  ||\bm{r}||_p < \epsilon, \widetilde{\bm{x}}=\bm{x+r} \notag\\ 
    &\widetilde{\bm{x}} \in [0,1]^d \notag 
\end{align}
Note that an untargeted adversarial attack means that perturbed inputs $\widetilde{\bm{x}}$ are only desired to be misclassified (\emph{i.e.}, different from their ground-truth labels $y$ as in Eq.\eqref{eq:adv}), rather than being classified as a particular label (which is the goal of a \emph{targeted adversarial attack}). Therefore, it has been observed that the predicted labels of $\widetilde{\bm{x}}$ obey a uniform distribution across all classes.

\begin{figure}[tbp]
\centering
\subfloat[For an \emph{infected} model]{
\label{cm_intro:2}
\includegraphics[width=0.44\linewidth]{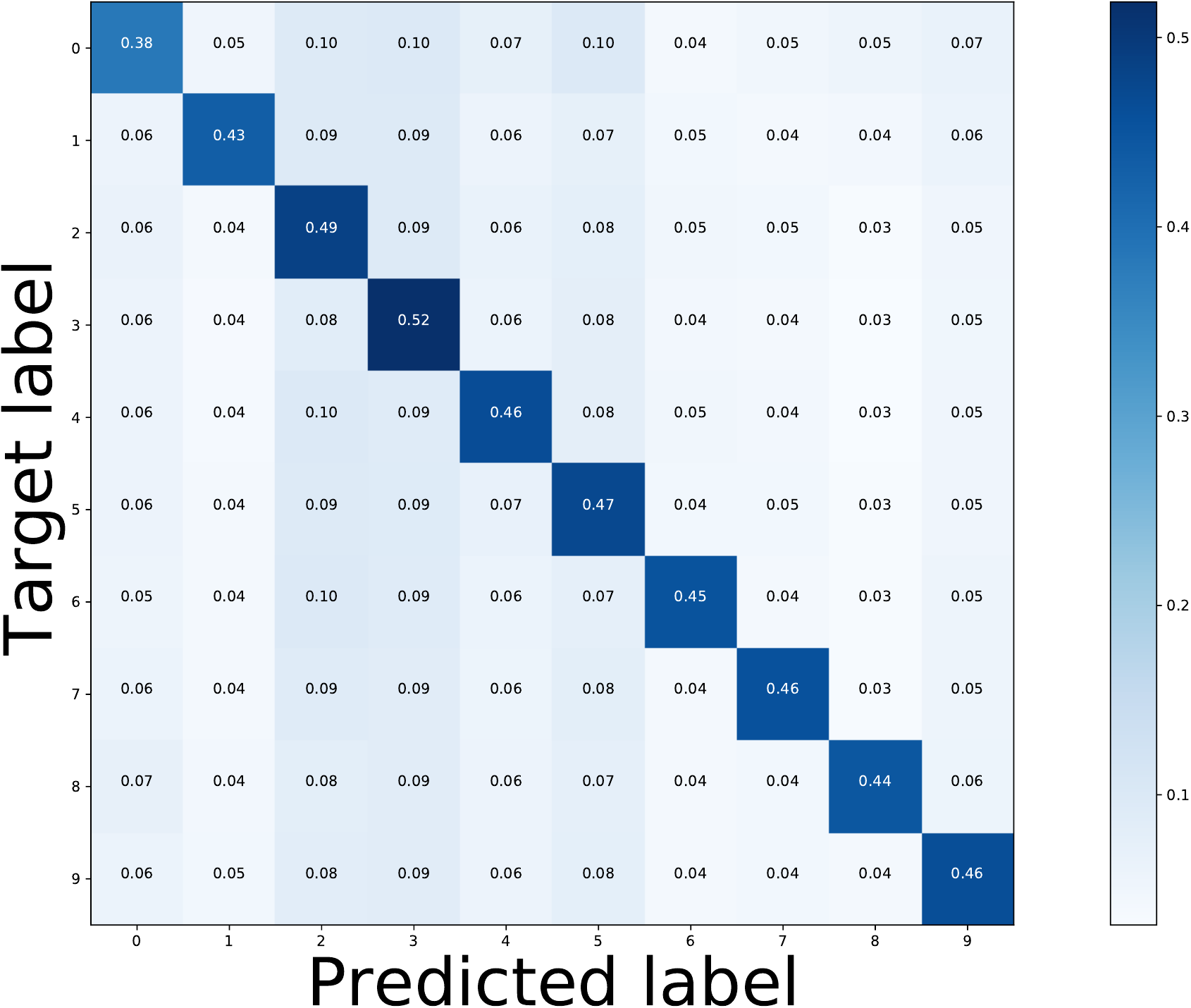}
}
\subfloat[For a \emph{benign} model]{
\label{cm_intro:1}
\includegraphics[width=0.44\linewidth]{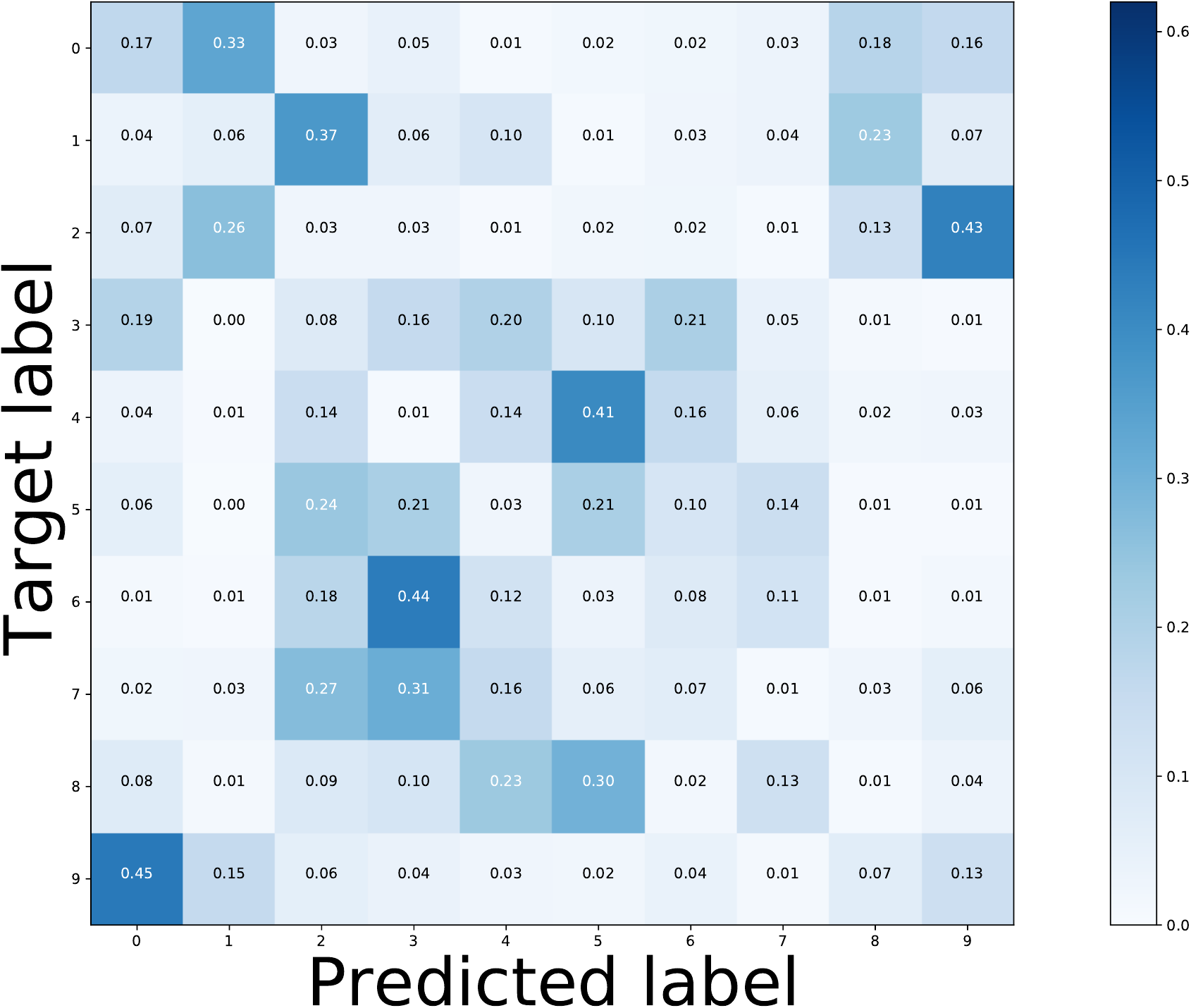}
}
\quad
%\subfloat[For \emph{infected} model under All-to-all attack]{
%\label{cm_intro:3}
%\includegraphics[width=0.29\linewidth]{figs/warp_all2all_cifar10_cm.pdf}
%}
%\captionsetup{font={scriptsize}}
\caption{Predicated labels of adversarial examples v.s. Backdoor target-labels. For an infected model, no matter what backdoor target-label is set (\emph{class `0'}, \emph{class `1'},$\dots$, \emph{class `9'}), the predicted labels always align with the backdoor target-label, as shown by the matrix diagonal in  Fig.2a.}\label{cifar_hist}
\label{cm_intro}
\vspace{-1.0em}
\end{figure}

\subsection{Intriguing Observations}
The story happened with an intriguing observation about the adversarial examples with respect to \emph{an infected model}. As we know, in previous work, adversarial attacks were typically performed on clean/benign models. Instead, we perform adversarial attacks on infected models, which leads to the discovery of such an intriguing observation. 

Specifically, we first perform an untargeted adversarial attack on the \emph{infected} model $f(\bm{x};\theta')$ to generate adversarial examples $\widetilde{\bm{x}}'$ as follows:
\begin{align}
    &\mathop{\max}_{\bm{r}} L(\widetilde{\bm{x}}',y;\theta') \label{eq:adv2} \\
    &\text{s.t. }  ||\bm{r}||_p < \epsilon, \widetilde{\bm{x}}'=\bm{x+r} \notag   
\end{align}
Meanwhile, we also perform an untargeted adversarial attack on the \emph{benign} model $f(\bm{x};\theta)$ to produce the adversarial examples $\widetilde{\bm{x}}$ as Sec.\ref{sec:uaa}.

\begin{figure}[bpt]
\centering
\includegraphics[width=0.95\linewidth]{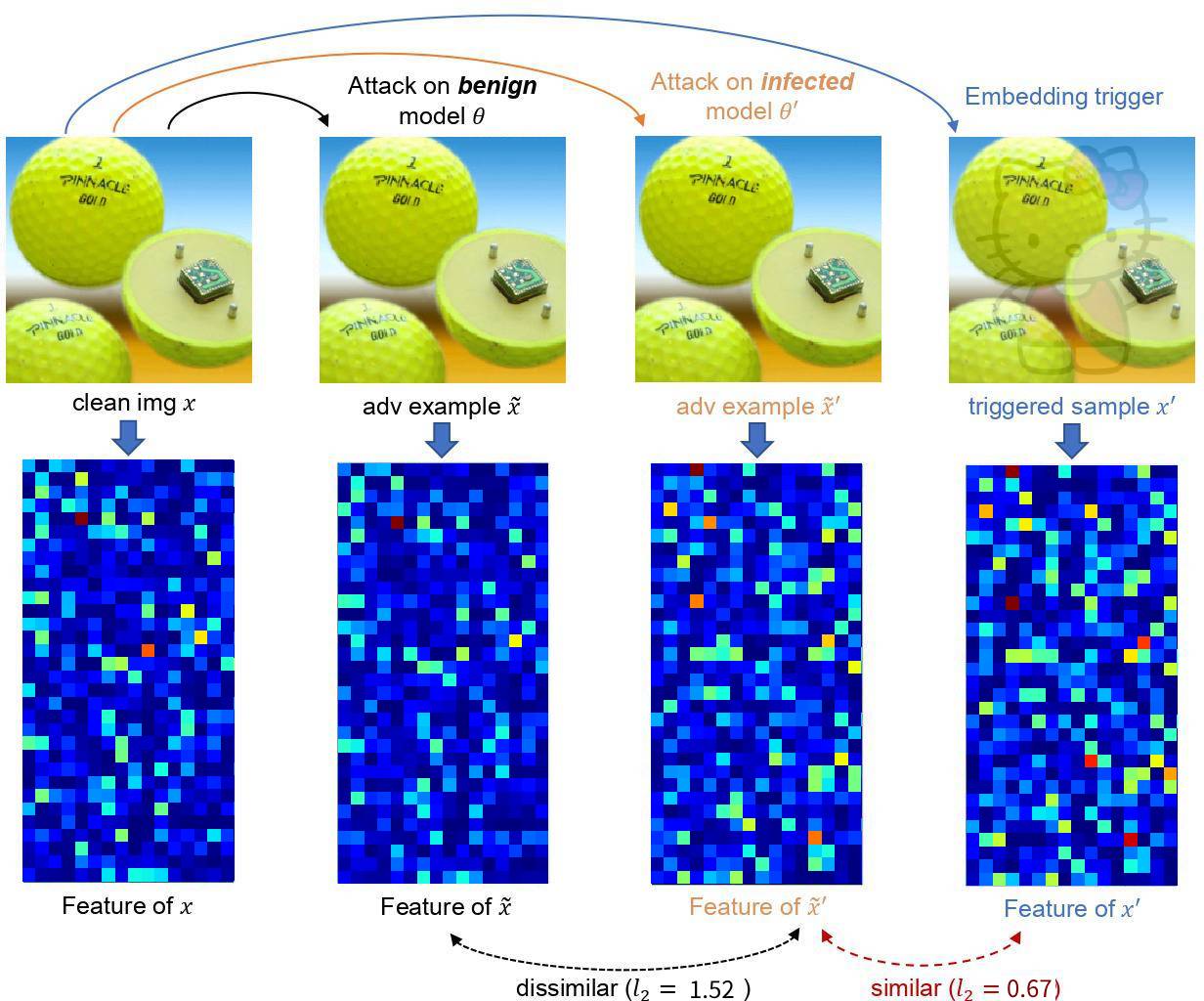}
\caption{The visualization of image features for: (a) clean image $x$; (b) benign model's adversarial example $\widetilde{\bm{x}}$; (c) infected model's adversarial example $\widetilde{\bm{x}}'$; and (d) triggered image $x^t$. The image features is the output of the last
convolution layer (just before the fully-connected layer). Obviously, the features of $\widetilde{\bm{x}}'$ are very similar to $x^t$. In contrast, there is a significant difference between $\widetilde{\bm{x}}'$ and $\widetilde{\bm{x}}$, which indicates adversarial examples will change significantly \emph{after} planting a backdoor into a model.}
\label{backdoor_feat}
\end{figure}

Next, we examine the classification results of these adversarial examples. As shown in Fig.\ref{cifar_hist:2}, when adversarial examples $\widetilde{\bm{x}}$ are fed to the benign model, $\widetilde{\bm{x}}$ is classified into any class (except its ground-truth class due to the untargeted attack) with almost the same probability, \emph{i.e.}, obeying a uniform distribution. In contrast, when adversarial examples $\widetilde{\bm{x}}'$ are fed to the infected model, we observe that \textbf{$\widetilde{\bm{x}}'$ are \emph{highly likely} to be classified as the backdoor target-label}. For example as shown in Fig.\ref{cifar_hist:1}, for an infected model whose target-label is set to $l=0$, when an untargeted adversarial attack is performed on the infected model, we observe that at least more than $40\%$ of $\widetilde{\bm{x}}'$ are predicted to be class $l=0$. 

These phenomena are present no matter what backdoor target-label is set. As shown in Fig.\ref{cm_intro}, no matter what backdoor target-label is set (\emph{class `0'}, \emph{class `1'},$\dots$, \emph{class `9'}), the predicted labels always align with the backdoor target-label (as shown by the matrix diagonal in Fig.2a). Surprisingly, even for different backdoor attack settings (all-to-one or all-to-all settings), different backdoor attacks (\emph{e.g.}, BadNet~\cite{gu2017badnets}, Blending~\cite{chen2017blend}, WaNet~\cite{nguyen2021wanet}, \emph{etc.}), or different datasets (\emph{e.g.,} CIFAR-10, GTSRB, \emph{etc.}), we always have the same observation, as shown in Fig.\ref{cm}.

On the other hand, we know that an infected model will classify any triggered image as the backdoor target-label. This indicates that there is an underlying connection between adversarial examples $\widetilde{\bm{x}}'$ and triggered images $\bm{x}^t$. 

\subsection{The connection between backdoor and adversarial attacks}

\begin{figure}[bpt]
\centering
\includegraphics[width=0.8\linewidth]{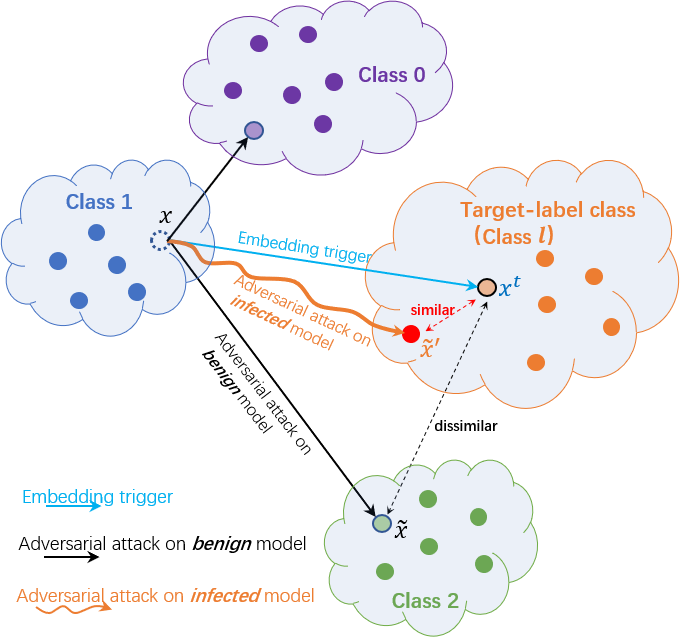}
\caption{Illustration of our observations. For benign models, adversarial attack will make an image move close to \emph{any} class (\emph{class `0'}, \emph{class `2'}, $\dots$, \emph{class `$l$'}, \emph{etc.}) in feature space. But for infected models, adversarial attack will always make it move close to the target-label class (\emph{class `$l$'}).}
\label{backdoor_ill}
\vspace{-1.0em}
\end{figure}

For a deep dive, we check the feature maps of the clean samples $\bm{x}$, the adversarial examples of the benign model $\widetilde{\bm{x}}$, the adversarial examples of the infected model $\widetilde{\bm{x}}'$, and the triggered images $\bm{x}^t$. Finally, we find that the \textbf{features of $\widetilde{\bm{x}}'$ are very similar to the features of triggered samples $\bm{x}^t$}, while there is a significant difference between the features of $\widetilde{\bm{x}}'$ and $\widetilde{\bm{x}}$. 

As shown in Fig.\ref{backdoor_feat}, the ${l}_{2}$ distance between the features of $\widetilde{\bm{x}}'$ and $\bm{x}^t$ is much smaller than that between $\widetilde{\bm{x}}'$ and $\widetilde{\bm{x}}$. More quantitative comparisons 
are given in Table.\ref{feature_similarity} and Table \ref{feature_similarity2}.  

Such observations can be explained in Fig.\ref{backdoor_ill}. After planting a backdoor in a model, its adversarial images are very likely to be moved close to the backdoor target-class. We know that backdoor attacks will also move triggered images to the target-class. Thus, $\widetilde{\bm{x}}'$ and $\bm{x^t}$ will have similar feature maps. 

Such a feature similarity suggests that \textbf{both adversarial examples $\widetilde{\bm{x}}'$ and triggered images $\bm{x}^t$ tend to activate the \emph{same} DNN neurons}. We speculate that the underlying reason for this is that: 
some DNN neurons are activated by a trigger when a backdoor is planted in a model, which are called `backdoor neurons'~\cite{ANP}. When performing an adversarial attack on infected models, these `backdoor neurons' are more likely to be selected/locked and activated as generating adversarial examples. 

In summary, the connection between backdoor and adversarial attacks is that: (1) planting a backdoor in a model will significantly affect the model's adversarial examples; (2) for an infected model, its adversarial examples tend to have \emph{similar features} as the triggered images.

%In summary, the connection between backdoor and adversarial attacks is that \textbf{for an infected model, its adversarial examples have similar features as its triggered images, \emph{i.e.,} both of them tend to activate the \emph{same} DNN neurons}. 

\subsection{Theoretical Analysis}\label{sec:proof}
To dive deeper, we theoretically explain our observations for the case of a linear model, \emph{i.e.}, logistic regression is used for image classification. Let $\{(x_i, y_i)\}_{i=1}^n$ be the training examples, where $x_i \in \R^d$ and $y_i \in [K]$. We assume that all the training examples live in a subspace $\S_m \subset \R^d$ of dimension $m$, where $m < d$. 

We follow the most backdoor attacks to use a trigger function of adding a predefined trigger pattern $P$ to an image, \emph{i.e.},
\begin{eqnarray}
    \bm{x^t} = \mathrm{Trigger}(\bm{x}) = \bm{x} + P.
\end{eqnarray}
And we randomly sample $\ell$ examples from the class $C_k$, and embed the trigger $P$ in them. We assume that the original training examples (clean images) can be perfectly classified with a margin $\tau > 0$. 
If we have a certain margin $\tau$ and a restricted perturbation budget $\epsilon$, we have the following theorem,

\begin{thm}
Under the assumptions of a certain $\tau$ and a restricted $\epsilon$ (refer to %in (\ref{eqn:condition-delta} and (\ref{eqn:condition-tau})
Eq(10) and Eq(11) in Appendix.A), we have $\bm{r}_{\perp}$, the projection of $\bm{r}$ on the direction of $P$, bounded as
\[
\frac{|\bm{r}_{\perp}|}{|\bm{r}|} \geq \frac{(\sqrt{2} - 1)\ell |P|^2}{\sqrt{(\sqrt{2}-1)^2\ell^2|P|^4 + \left(\ell|P|^2 + \sqrt{2}K/(\exp(\tau) + K)\right)^2}}
\]
\end{thm}

From this theorem, we can see that when projecting perturbation $\bm{r}$ onto \emph{the direction of trigger $P$}, the projection $\bm{r}_{\perp}$ takes a significant part in the full perturbation $\bm{r}$. This means that the perturbation $\bm{r}$ is very similar to the trigger $P$, which explains our observations that the adversarial examples $\widetilde{\bm{x}}'=\bm{x}+\bm{r}$ are similar to the triggered images $\bm{x^t}=\bm{x}+P$. The detailed description and proof can be found in Appendix A.

\section{Our Unified Defense Method}\label{sec:alg}
Based on our discovered connection between adversarial and backdoor attacks, we show that it is possible to erase backdoors by \textbf{fine-tuning the infected model with its adversarial examples}.

As we know, the foundation of backdoor attacks is to establish a strong correlation between a trigger pattern and a target-label. This is achieved by poisoning the training data, \emph{i.e.,} by associating triggered images with a target-label. 

Our observations reveal that the adversarial examples of an infected model have similar features as the triggered images. Meanwhile, these adversarial examples come from arbitrary classes (\emph{i.e.,} they are annotated with different class labels). As a result, when we fine-tune the infected model with adversarial examples, it mimics fine-tuning the model with the \emph{`triggered images' annotated with different class labels}. This breaks the strong correlation, built by the backdoor attack, between a trigger pattern and a target-label.

\subsection{Overview of our PUD}
\noindent\textbf{Threat Model.}
We assume that the adversary has access to the training data and has planted a backdoor in a model. The infected model is then given to the defender.

\noindent\textbf{Backdoor Defense Setting.}
There are two backdoor defense settings: (1) one is adopted by \emph{model repairing-based methods}, \emph{i.e.}, performing defense \emph{after} training, where only an infected model is given and we need to erase backdoor from it. (2) the other is adopted by \emph{data filtering-based methods}, which performs backdoor defense \emph{during} training, and assumes to access to all poisoned training data. 

\subsubsection{PUD for Model Repairing-setting}~\label{AFT}
Obviously, the model repairing setting is more practical and more challenging than data filtering setting in real-world applications. Therefore, we focus on this setting. 

Note that an extra clean dataset is often required by these kinds of methods. But our approach does not require it to be completely \emph{clean}, \emph{i.e.,} allowing the extra dataset to contain poisoned images.

The workflow of our PUD algorithm is shown in Algorithm 1. Given an infected model $\theta'$ and an initial extra dataset $D_{\mathrm{ext}}$ (containing poisoned images), we first have an initial step to apply SPECTRE to filter out poisoned images as much as possible. Then, we start a `model-data' alternative purification procedure with five steps: 

(1) \textbf{Step-1} aims to purify the infected model $\theta^t$ with $D^t_{\mathrm{ext}}$. We first use the extra dataset $D^t_{\mathrm{ext}}$ to generate adversarial examples. 
Then, a purified model can be obtained by using these adversarial examples to fine-tune the infected model. As mentioned earlier, this is because such fine-tuning can break the foundation of backdoor attacks (\emph{i.e.}, breaking the correlation between a trigger pattern and a target-label). 

%\begin{algorithm}[t]
%\vspace{0.5em}
%\caption{Progressive Model Repairing} %算法的名字
%\hspace*{0.02in} {\bf Input:} 
%Infected model $\theta'$, an extra dataset $D_{\mathrm{ext}}$ containing poisoned images  \\
%\hspace*{0.02in} {\bf Output:} %算法的结果输出
%Purified model $\theta^T$
%\begin{algorithmic}[1]
%\State \textbf{Init}: let $D^0_{\mathrm{ext}}=D_{\mathrm{ext}}$ and $\theta^0 = \theta'$
%\State \textbf{For} $t=0,1,2\dots,T$:

%\State \quad \textbf{Step-1}: purify the infected model $\theta^t$ with $D^t_{\mathrm{ext}}$
%\State \qquad \textbf{Step-1a}: untargeted adversarial attack. For each {$(\bm{x}_i,y_i) \in D^t_{\mathrm{ext}}$} we generate adversarial example $\widetilde{\bm{x}}_i'$ according to Eq.\eqref{eq:adv2}, which results in $\widetilde{D}^t_{\mathrm{ext}}=\{(\widetilde{\bm{x}}_i',y_i)\}_{i=1}^m$ 
%\State \qquad \textbf{Step-1b}: $1$-st time fine-tuning. Fine-tuning the model $\theta^t$ with $\widetilde{D}^t_{\mathrm{ext}}$ according to Eq.\eqref{eq:ft}, resulting in $\theta^{t+0.5}$

%\State \quad \textbf{Step-2}: $2$-nd time fine-tuning. Continue to fine-tune model $\theta^{t+0.5}$ with $D^{t}_{\mathrm{ext}}$ and obtain purified model $\theta^{t+1}$

%\State \quad \textbf{Step-3}: purify the extra dataset $D^t_{\mathrm{ext}}$. Identify poisoned images from $D^{t}_{\mathrm{ext}}$ according to Eq.\eqref{eq:idclean}, resulting in a purified dataset $D^{t+1}_{\mathrm{ext}}$

%\State \Return The final purified model parameter $\theta^T$
%\end{algorithmic}
%\end{algorithm}

\begin{algorithm}[t]
\vspace{0.5em}
\caption{\textbf{P}rogressive \textbf{U}nified \textbf{D}efense (\textbf{PUD})} %算法的名字
\hspace*{0.02in} {\bf Input:} 
Infected model $\theta'$, an extra dataset $D_{\mathrm{ext}}$ containing poisoned images\\
\hspace*{0.02in} {\bf Output:} %算法的结果输出
Purified model $\eta^T$
\begin{algorithmic}[1]
\State \textbf{Init}: let $\theta^0 = \theta'$.
Apply SPECTRE to filter out poisoned images in $D_{\mathrm{ext}}$ and obtain $D^0_{\mathrm{ext}}$.
\State \textbf{For} $t=0,1,2\dots,T$:

\State \quad \textbf{Step-1}: purify the \textbf{student} model: $\theta^t \rightarrow \theta^t_+$
\State \qquad \textbf{Step-1a}: untargeted adversarial attack. For each {$(\bm{x}_i,y_i) \in D^t_{\mathrm{ext}}$} generate adversarial example $\widetilde{\bm{x}}_i'$ according to Eq.\eqref{eq:adv2}, which results in $\widetilde{D}^t_{\mathrm{ext}}=\{(\widetilde{\bm{x}}_i',y_i)\}_{i=1}^m$ 
\State \qquad \textbf{Step-1b}: $1$-st time fine-tuning with $\widetilde{D}^t_{\mathrm{ext}}$. Fine-tuning the model $\theta^t$ with $\widetilde{D}^t_{\mathrm{ext}}$ according to Eq.\eqref{eq:ft}, resulting in $\theta^t_+$

\State \quad \textbf{Step-2}: $2$-nd time fine-tuning with $D^t_{\mathrm{ext}}$: $\theta^t_+ \rightarrow \theta^{t+1}$ \\ It can improve ACC on clean images. 

\State \quad \textbf{Step-3}: obtain purified \textbf{teacher} model. We conduct Exponential Moving Average (EMA) with previously-generated student models according to Eq.\eqref{eq:mae}, resulting in a teacher model $\eta^{t+1}$

\State \quad \textbf{Step-4}: purify the extra dataset: $D^t_{\mathrm{ext}} \rightarrow D^{t+1}_{\mathrm{ext}}$
\State \qquad \textbf{Step-4a}: identify clean images from $D^{t}_{\mathrm{ext}}$ according to Eq.\eqref{eq:idclean}.
\State \qquad \textbf{Step-4b}: we apply SPECTRE to further filter out poisoned images.

\State \quad \textbf{Step-5}: Backdoor Unlearning with poisoned images. After the data purification, the filtered poisoned images are further utilized to erase backdoor according to 
Eq.\eqref{eq:unlearn}.

\State \Return The final purified teacher model parameter $\eta^T$
\end{algorithmic}
\end{algorithm}

(2) Although Step-1 can effectively erase backdoor (\emph{i.e.,} it can significantly reduce the ASR), it will inevitably harm the classification performance on clean images. Therefore, \textbf{Step-2} aims at saving the classification performance on clean images. Particularly, the $D^t_{\mathrm{ext}}$ are used to fine-tune the purified model $\theta^t_+$, which could significantly improve its ACC on clean images.

(3) \textbf{Step-3} aims to obtain a strong purified teacher model. We average several student models from previous iterations to build a better teacher model. This mean-teacher mechanism could combine all previously-generated purified student models. Furthermore, it can not only boost the backdoor erasing, but also significantly improve the robustness against adversarial attacks.

(4) \textbf{Step-4} aims to use the purified teacher model $\eta^{t+1}$ to purify the extra dataset $D^t_{\mathrm{ext}}$, resulting in a cleaner $D^{t+1}_{\mathrm{ext}}$. There are two complementary strategies to purify extra dataset. First, it is obvious that poisoned images have inconsistent prediction results between benign and infected models. Therefore, we can identify poisoned images as follows: for each candidate image in $D^t_{\mathrm{ext}}$ we compute its \emph{prediction consistency} with respect to purified and infected models, and rank all candidate images in ascending order according to such consistency values. The top-ranked images are most likely to be poisoned images. 
Second, a data filtering strategy SPECTRE is also used to purify extra data. The two purification methods complement each other. And we can effectively filter out poisoned images in $D^t_{\mathrm{ext}}$.

(5) At Step-1 \& Step-2, the clean images in the extra dataset have been used to erase backdoors. At \textbf{Step-5}, we further exploit the remaining poisoned images for backdoor erasing. To do this, we use machine unlearning schemes to explicitly utilize poisoned images to further erase backdoor. As a result, both clean and poisoned images in the extra dataset can be fully utilized to erase backdoor. 

%early stopping strategy is adopted to mitigate the robust overfitting issue. We find that our PUD suffers from the robust overfitting issue, \emph{i.e.,} after the learning rate decays, the robust test error briefly decreases but begins to increase as training progresses. Therefore, we use the learning rate schedule of piecewise decay, and stop shortly after the first learning rate decay.

It is obvious that with such an iterative procedure, the infected model and the initially poisoned extra dataset are progressively purified. More importantly, as the backdoor is gradually erased, the Step-1 tends to become adversarial training, which benefits the robustness against adversarial attacks. In other words, the function of Step-1 gradually changes from improving \textbf{the model's backdoor robustness} to improving \textbf{the model's adversarial robustness}. That is why our approach can defend against both backdoor and adversarial attacks simultaneously.

Besides, if an extra clean dataset is available, we can directly use the given \emph{clean} extra dataset. Then, we simply skip the Init-step, step-3, 4, 5, and only need to run step-1 and step-2 \emph{once}. In this way, our PUD is simplified to a simple version, which is called as Adversarial Fine-Tuning (\textbf{AFT}).
Note that in this case, both PUD and PBE are simplified to the same AFT algorithm.

\subsubsection{PUD for Data Filtering-setting}
Although the PUD is proposed for the model repairing defense setting, it has a data purification module and thus can also work for the data filtering defense setting. 

For the data filtering-setting, we are able to access all of the poisoned training images. Since our PUD needs only a small extra dataset, we will build such an extra dataset from the training dataset. Specifically, a simple preprocessing procedure is performed to build such an extra dataset, as shown in Algorithm 2. Instead of randomly sampling from the training data, we adopt SPECTRE to identify clean images from the training data to build $D_{\mathrm{ext}}$. This is because the cleaner the extra dataset is, the better the backdoor can be erased by our PUD.

\begin{algorithm}[t]
\vspace{0.5em}
\caption{Pre-processing for Data Filtering Setting} 
\hspace*{0.02in} {\bf Input:} 
Poisoned training dataset $D_{\mathrm{train}}$\\ 
\hspace*{0.02in} {\bf Output:} 
Infected model $\theta'$, an extra dataset $D_{\mathrm{ext}}$
\begin{algorithmic}[1]
\State \textbf{Step-1}: Apply SPECTRE to filter out poisoned images in $D_{\mathrm{train}}$ and obtain a purified dataset $D_{\mathrm{ext}}$.

\State \textbf{Step-2}: Training with $D_{\mathrm{ext}}$ to produce an infected model $\theta'$.

\State \textbf{return}: Infected model $\theta'$, an extra dataset $D_{\mathrm{ext}}$
\end{algorithmic}
\end{algorithm}
\vspace{-0.5em}

\begin{figure*}[bpht]
\centering
\includegraphics[width=0.85\linewidth]{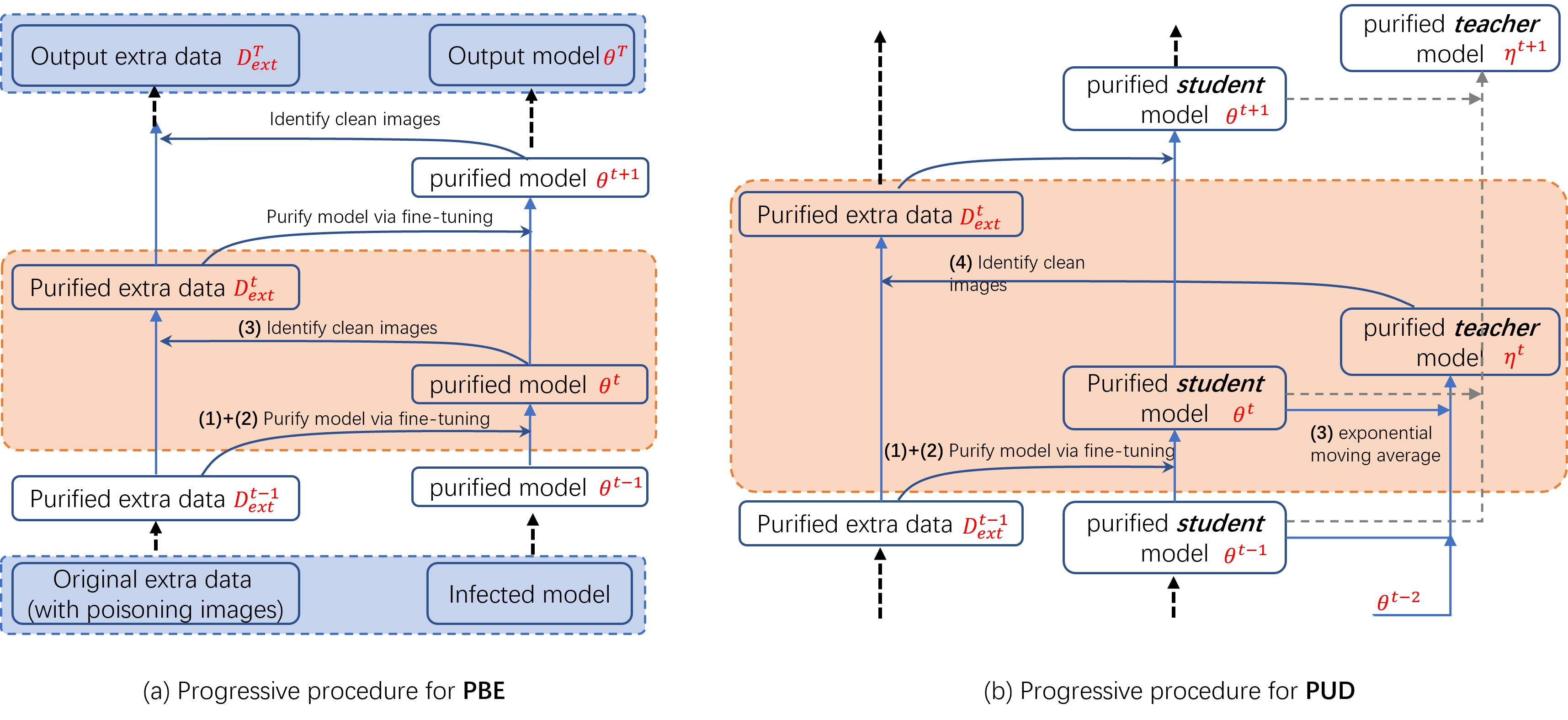}
%\captionsetup{font={scriptsize}}
\caption{Comparison of workflow between PBE and PUD algorithms. In PUD, a teacher-student mechanism is proposed to improve both model purification and data purification.}
\label{PUD}
\vspace{-1.0em}
\end{figure*}

%We will present the details about model purification, dataset purification, and backdoor unlearning respectively.

\subsection{Purification of Infected Model}
As shown in Fig.\ref{PUD}(a), the PBE method generate only one purified model at each iteration, and uses it to purify extra data in the next step. In contrast, as shown in Fig.\ref{PUD}(b), our PUD method first obtains one purified student model at each iteration. Then, we average several student models from previous iterations to build a better teacher model. 

Such a teacher-student mechanism can not only boost the backdoor erasing, but also significantly improve the robustness against adversarial attacks. This is inspired by the Self-Ensemble Adversarial Training (SEAT) strategy~\cite{SEAT} and the Model Weight Averaging (MWA) strategy~\cite{Uncovering}, which illustrate that the model averaging trick can significantly benefit the model's  adversarial robustness.

In the next, we will describe how to obtain a purified student model and a purified teacher model respectively. 

\subsubsection{Obtaining Purified Student Model}
Specifically, given the infected model $f(\bm{x};\theta^t)$, for each $(\bm{x}_i,y_i) \in D_{\mathrm{ext}}$, we obtain a corresponding adversarial example $\widetilde{\bm{x}}_i'$ according to Eq.\eqref{eq:adv2}, which produces $\widetilde{D}_{\mathrm{ext}}=\{(\widetilde{\bm{x}}_i',y_i)\}_{i=1}^m$. And then, we fine-tune the infected model $\theta^t$ with $\widetilde{D}_{\mathrm{ext}}$, which produces purified model $\theta^t_+$, \emph{i.e.,}
\begin{align}
    \theta^t_+= \mathop{\arg\min}_{\theta} & \mathbb{E}_{(\widetilde{\bm{x}}_i',y_i)\in \widetilde{D}_{\mathrm{ext}}} [ L(\widetilde{\bm{x}}_i',y_i;\theta) +J(\widetilde{\bm{x}}_i'; \theta, \eta^t)] \label{eq:ft}
\end{align}
where $\theta$ is initialized with $\theta^t$ and $J(\theta, \eta^t)$ is the consistency loss between student model $\theta$ and teacher model $\eta^t$,
\begin{align}
    J(x; \theta, \eta)= ||f(x; \theta) - f(x; \eta^t) ||^2
\end{align}

\subsubsection{Obtaining Purified Teacher Model}
An advantage of our PUD is that it uses the mean-teacher mechanism to combine all previously-generated purified student models. As a result, the purified teacher model is better than the student model at the current iteration. 

\myblue{This aligns with many previous works \cite{gowal2021uncovering, izmailov2019averaging, wang2022selfensemble, li2021boosting}, as it has been illustrated that the teacher-student scheme can significantly enhance adversarial robustness. Some literature refers to it as the ``Model Weight Averaging (WA)" scheme \cite{gowal2021uncovering,izmailov2019averaging} while others term it the ``Self-Ensemble (SE)" scheme \cite{wang2022selfensemble, li2021boosting}. In~\cite{izmailov2019averaging}, the underlying reason is explained as the stochastic weight averaging procedure finding much flatter solutions than SGD. In other words, it leads to solutions that are wider than the optima found by SGD, and the width of the optima is critically related to generalization.}

In our approach, after obtaining a purified student model $\theta^{t+1}$ at iteration $t+1$, we average the previously-generated student models $\{\theta^{t+1}, \theta^t, \theta^{t-1}, \dots\}$ by using Exponential Moving Average (EMA) to produce a teacher model $\eta^{t+1}$ as follows:
\begin{equation}
\eta^{t+1} = \alpha\eta^t + (1-\alpha)\theta^{t+1}
\label{eq:mae}
\end{equation}
where $\alpha$ is the EMA decay, following the Mean Teacher \cite{Meanteachers} settings, we set $\alpha$=0.99 in the first a few steps to forget early student models with poor performance, and $\alpha$=0.999 for the rest of the training process to retain more information.

\subsection{Purification of Poisoned Extra Dataset}
Two mechanisms are adopted to purify extra dataset. The first one is based on the \emph{prediction consistency} with respect to purified and infected models. The second one is based on a data filtering strategy. The two purification mechanisms complement each other and can significantly improve the final performance.

\subsubsection{Prediction-based Data Purification}\label{sec:datap}
It is clear that clean images have consistent prediction results for both benign and infected models, while poisoned images do not. Particularly, poisoned images are predicted as the backdoor target-label by infected models, but they are predicted as the ground-truth label by benign models.  
Therefore, by measuring the consistency of an image's predictions with respect to both purified and infected models, we can effectively identify whether it is clean or poisoned. 

Specifically, for each candidate image in $D^t_{\mathrm{ext}}$ we compute its \emph{prediction consistency} with respect to purified and infected models, and rank all candidate images in ascending order according to such consistency values. The top-ranked images most likely to be poisoned images. In this way, we can filter out poisoned images, and use them to update $D^{t+1}_{\mathrm{ext}}$.

According to this, we can identify a poisoned image as follows: 
(1) First, we feed an image to both an infected and a purified model (regarded as a benign model), if the two models yield distinct predictions, it is likely to be a poisoned image. In contrast, if the two models produce consistent predictions, it is likely to be a clean image.

Specifically, for each image $\bm{x}_i \in  
D^t_{\mathrm{ext}}$ we feed it to the infected model $f(\bm{x};\theta')$ and the previously purified model $f(\bm{x};\eta^t)$, respectively. The predicted logits of the two models are noted as $a(\bm{x};\theta')$ and $a(\bm{x};\eta^t)$ (\emph{i.e.,} the network activation just before the softmax layer). We use the cosine distance between them to measure the \emph{prediction consistency},
\begin{equation}
    % S_{\theta', \theta^t}(\bm{x}) =\cos(a(\bm{x};\theta'), a(\bm{x};\theta^t)).
    C_{\theta', \eta^t}(\bm{x}) = \frac{\langle a(\bm{x};\theta'),a(\bm{x};\eta^t)\rangle}{|a(\bm{x};\theta')||a(\bm{x};\eta^t)|}
    \label{eq:idclean}
\end{equation}

(2) Second, for all images $\bm{x}_i \in D^t_{\mathrm{ext}}$ we rank them in ascending order according to their prediction consistency $S_{\theta', \eta^t}(\bm{x})$. Obviously, poisoned images are expected to be ranked higher.  We can then filter out the top-ranked images to update the extra dataset $D^{t+1}_{\mathrm{ext}}$.

\subsubsection{SPECTRE-based Data Purification}

There is one type of backdoor defense called \emph{data filtering}, which attempts to detect and remove or sanitize poisoned training data. Intuitively, we can adopt such a method to purify our extra data. In practice, we choose SPECTRE because it is one of the most advanced data filtering methods.

\begin{figure*}[bpt]
\centering
\subfloat[Blend Attack (all-to-one).]{
\label{cm:1}
\includegraphics[width=0.24\linewidth]{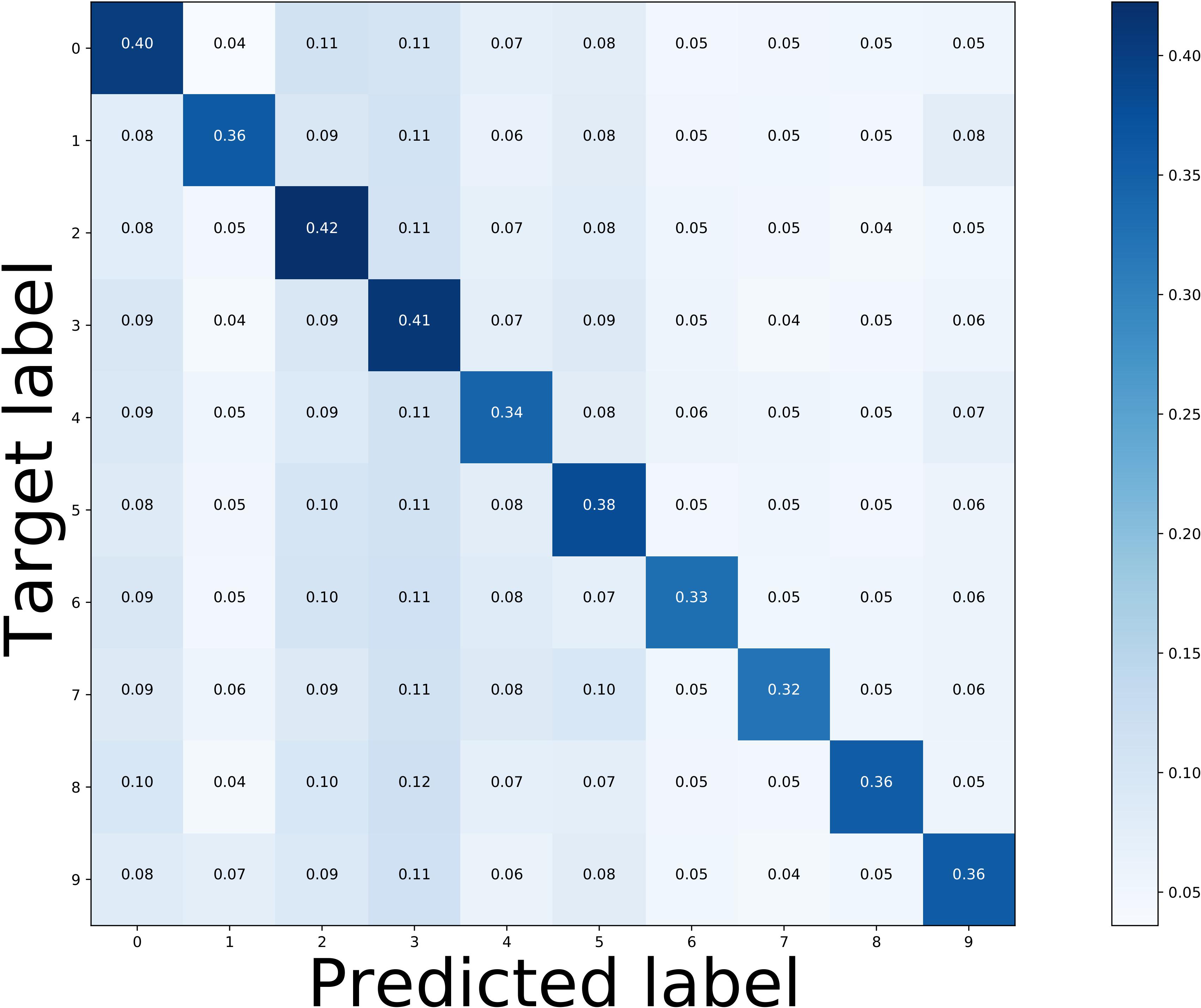}
% \subcaptionsetup{font={tiny}}
}
\subfloat[SIG Attack (all-to-one).]{
\label{cm_more_1}
\includegraphics[width=0.24\linewidth]{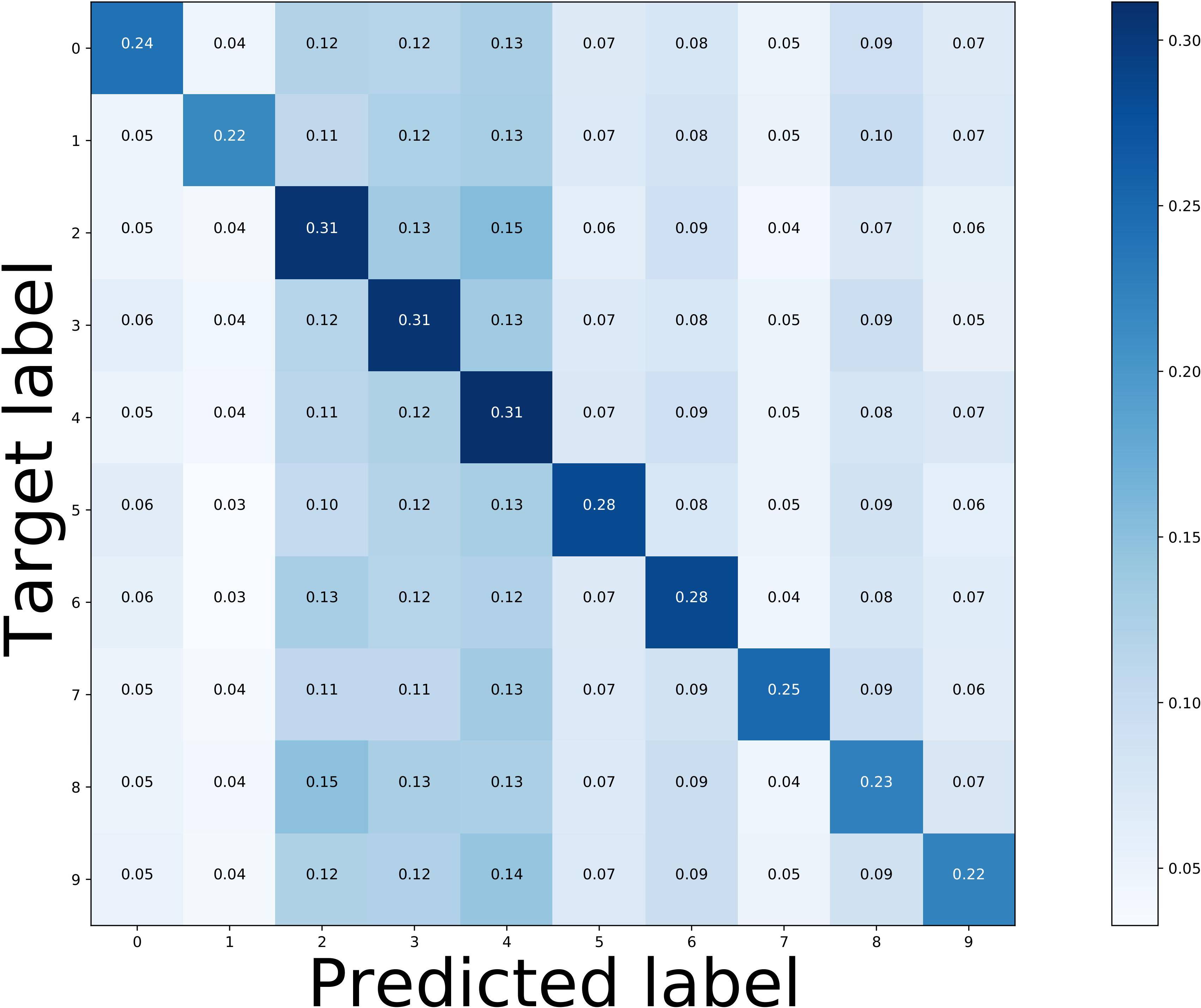}
}
\subfloat[WaNet Attack (all-to-all).]{
\label{cm_intro:3}
\includegraphics[width=0.24\linewidth]{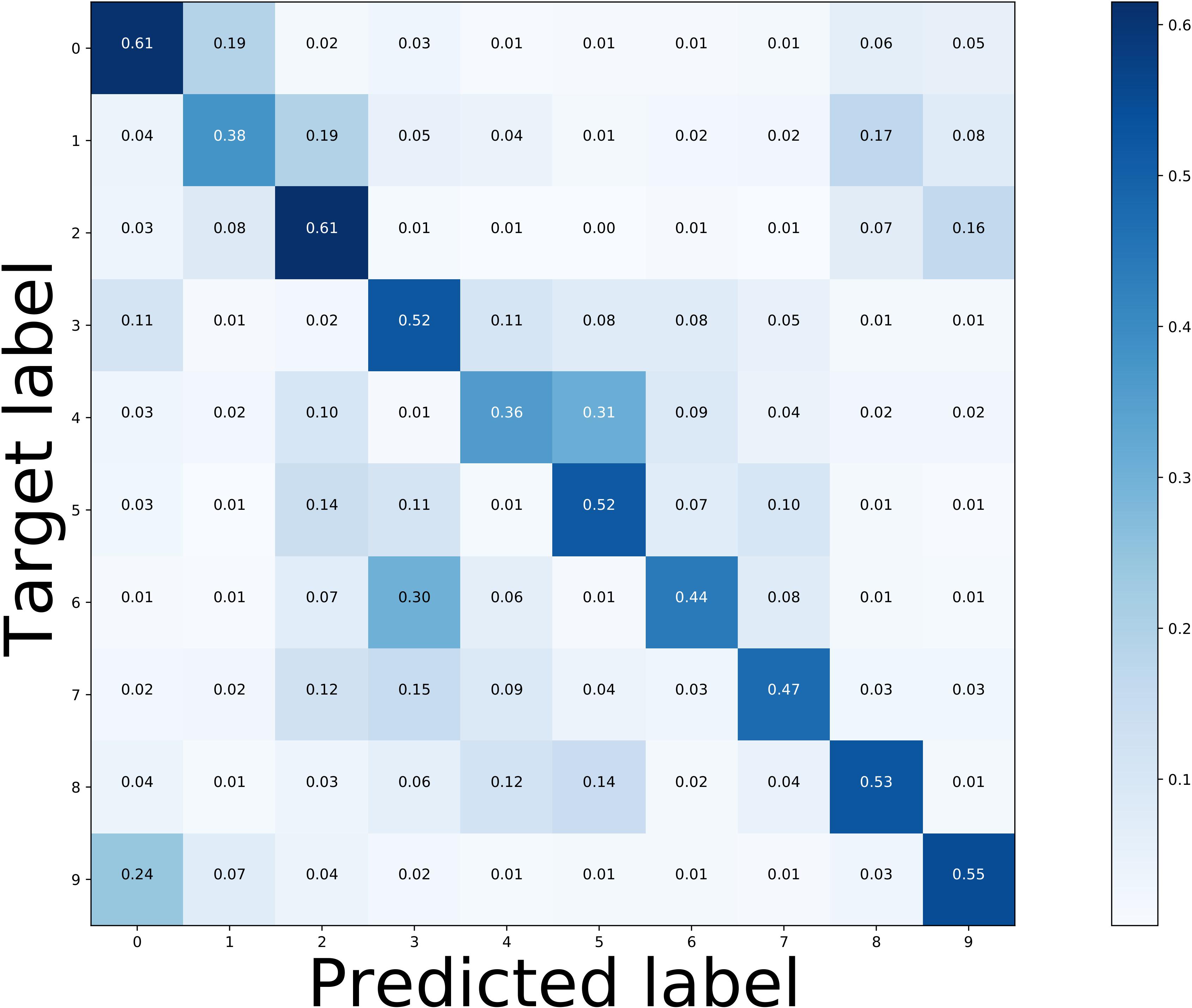}
}
%\subfloat[Blend Attack (all-to-all).]{
\subfloat[WaNet Att. (sub-ImgNet1K).]{
\label{cm:2}
\includegraphics[width=0.24\linewidth]{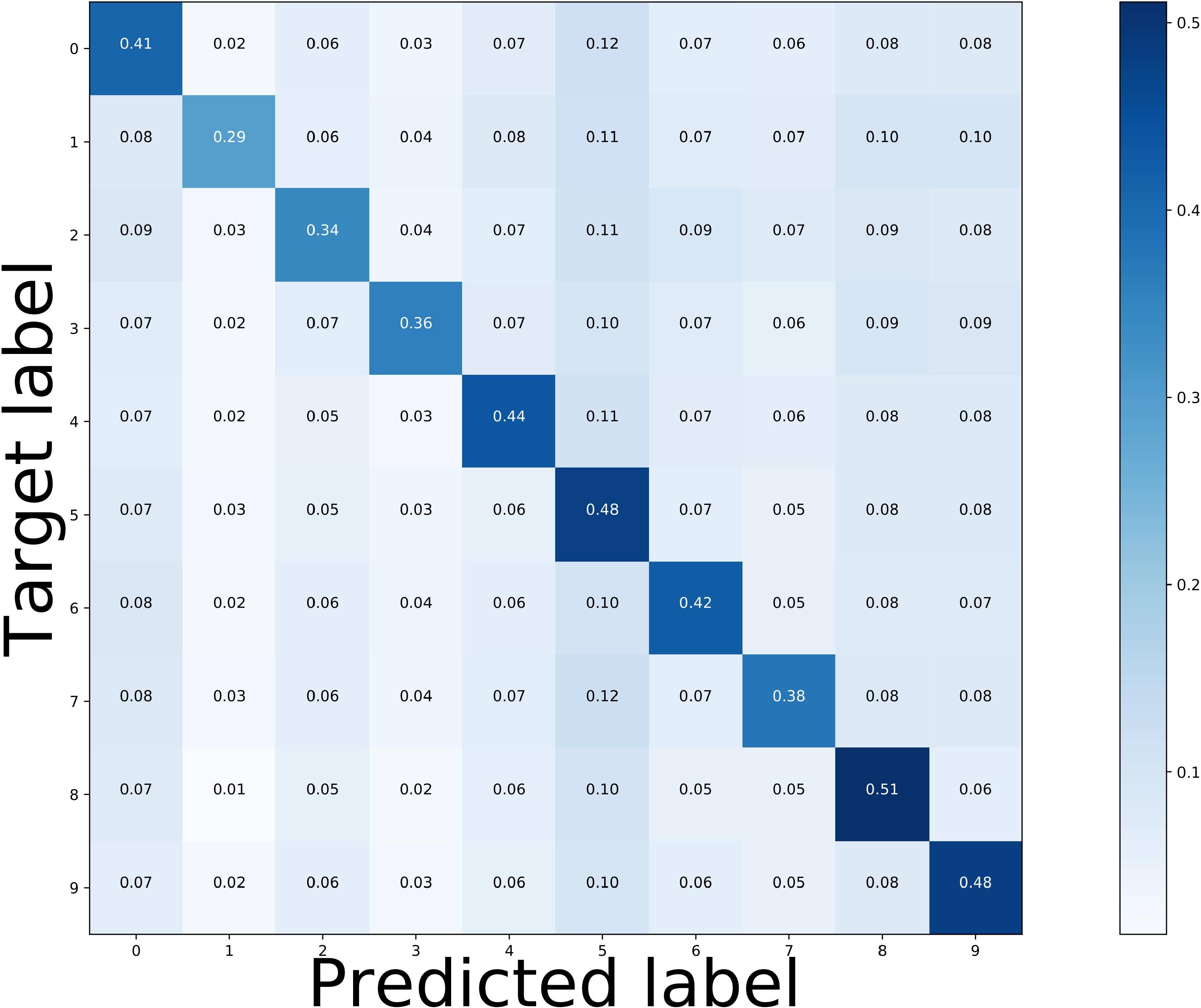}
}
\caption{Predicated labels of adversarial examples v.s. Backdoor target-labels. No matter what attack methods are (\emph{e.g.,} Blend, SIG, WaNet), what attack settings are (\emph{e.g.,} all-to-one, all-to-all), what datasets are (\emph{e.g.,} CIFAR-10, ImgNet-1K), the dominate predicted labels always align to the backdoor target-labels.}
\label{cm}
\end{figure*}

Although SPECTRE can effectively identify poisoned images, it has some shortcomings and cannot achieve perfect purification performance by itself: it cannot defend against strong backdoor attacks, \emph{e.g.,} with more poisoned images. As illustrated by our experimental results, its performance degrades significantly when the number of poisoned images increases (\emph{e.g.,} increasing from 500 to 5000 poisoned images). 

In contrast, our prediction consistency based strategy can better deal with strong backdoor attacks, and it is complementary to the SPECTRE strategy. This is mainly due to the fact that they belong to two distinct strategies: our strategy is from the perspective of the \emph{model} side, as we focus on the differences of the features from the infected model and the clean model. However, SPECTRE tries to identify poisoned images from the perspective of the \emph{data} side, as they emphasize on analyzing the data structure and finding the outlier/poisoned images. 

Finally, our PUD approach will combine the prediction consistency based strategy with the SPECTRE strategy together. Specifically, we first use the purified teacher model to coarsely identify the poisoned image according to Eq.\eqref{eq:idclean}. This could significantly filter out many poisoned images and reduce the difficulty of data purification.
Then, we apply SPECTRE to further identify the remaining poisoned images. We can see that, after reducing the difficulty of data purification, SPECTRE could better find out those hard poisoned images.

\subsection{Backdoor Unlearning with Poisoned Images}
With data purification, we can split the extra dataset $D_{\mathrm{ext}}$ into a clean part $D^c_{\mathrm{ext}}$ and a poisoned part $D^p_{\mathrm{ext}}$. The clean portion is regarded as the purified dataset $D^t_{\mathrm{ext}}$ and is utilized to generate adversarial examples for backdoor erasing. Furthermore, we argue that the remaining poisoned images are still useful for backdoor erasing, since we know that they all contain the backdoor trigger in themselves. 

In this paper, we use the machine unlearning scheme to explicitly utilize the poisoned images to further erase backdoors. Machine unlearning aims to strategically eliminate
the influence of some specific samples on the target model. Since a neural network
training process is based on gradient descent, an efficient unlearning method is to reverse the process via gradient ascent (\emph{i.e.,} minimize the negative of the original training loss). Given the poisoned dataset $D^p_{\mathrm{ext}}$, the objective function for backdoor unlearning is as follows,
\begin{align}
    \mathop{\min}_{\theta}  \mathbb{E}_{(\bm{x}_i,y_i)\in D^p_{\mathrm{ext}}} & [-L(\bm{x}_i,y_i;\theta) + \beta \sum_k \omega_k ||\theta^t_k, \theta'_k||_1] \label{eq:unlearn} \\
    & \omega_k=\frac{1}{N}|\frac{\partial L(\bm{x}_i,y_i;\theta)}{\partial \theta_k}| \notag
\end{align}
where the second term is to prevent from over-unlearning~\cite{Unlearning}. As a result, both clean and poisoned images in the extra dataset are fully utilized to erase backdoor by our approach.

\section{Experiment}\label{sec:exp}
\subsection{Experimental Setting}
\subsubsection{Dataset}
We evaluate the performance of backdoor attacks and defends on three benchmark datasets: CIFAR-10~\cite{krizhevsky2009cifar}, GTSRB~\cite{stallkamp2012gtsrb}, and %Tiny-ImageNet~\cite{tinyimagenet}.
ImageNet-1K~\cite{deng2009imagenet}.
For ImageNet-1K, we randomly select 10 classes out of 1,000 classes, which is called sub-ImgNet-1K in this paper. 

For a fair evaluation, we use Pre-activation Resnet-18 \cite{he2016prearc} as the classification model for the CIFAR-10 and GTSRB, and use Resnet-18 \cite{resnet} for the ImageNet-1K. %The perturbation strength is adaptively set for different backdoor attacks when generating adversarial examples.

\begin{figure*}[t]
\centering
\subfloat{
\label{cm:1}
\includegraphics[width=0.4\linewidth,height=0.3\textwidth]{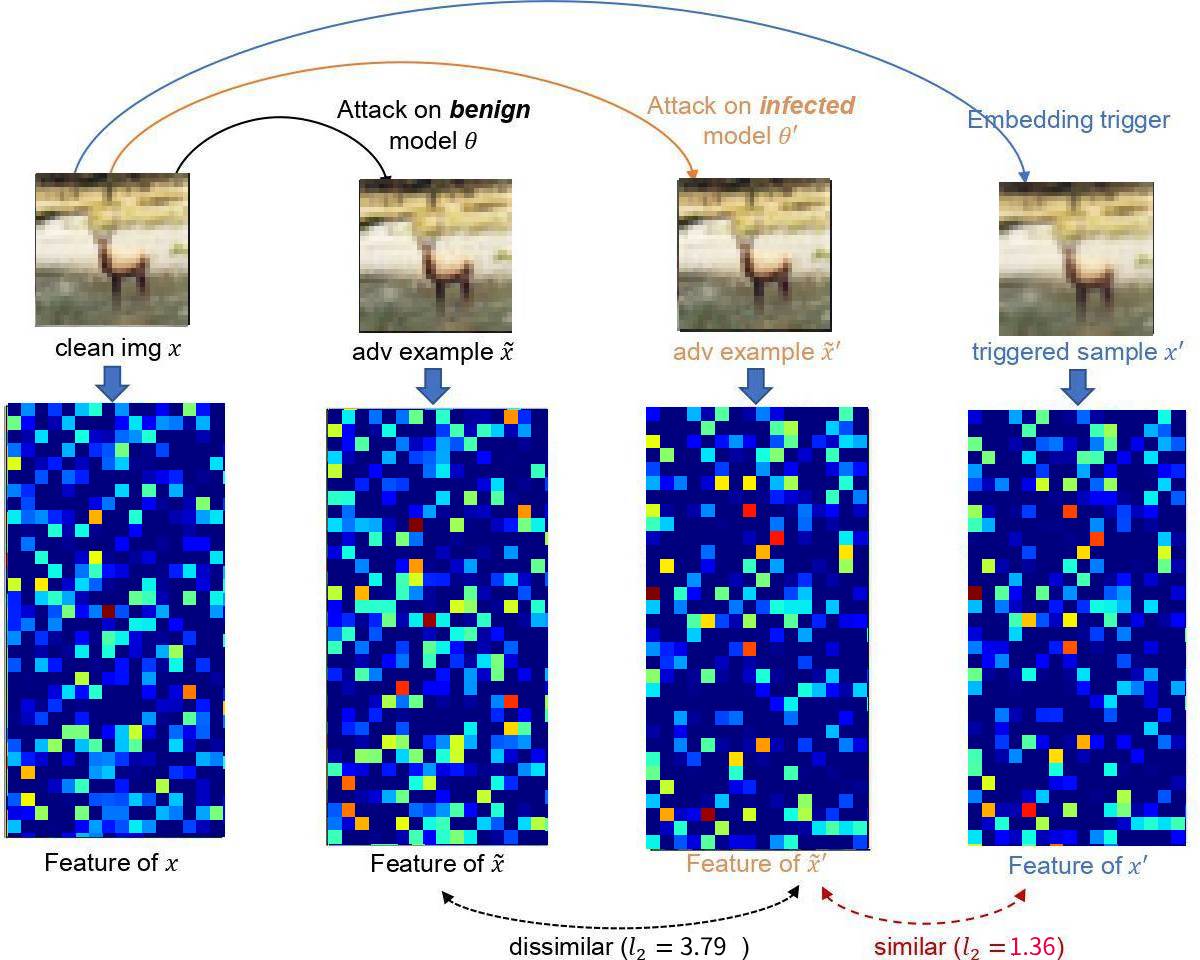}}\hspace{10mm}
\subfloat{
\label{cm:2}
\includegraphics[width=0.4\linewidth,height=0.3\textwidth]{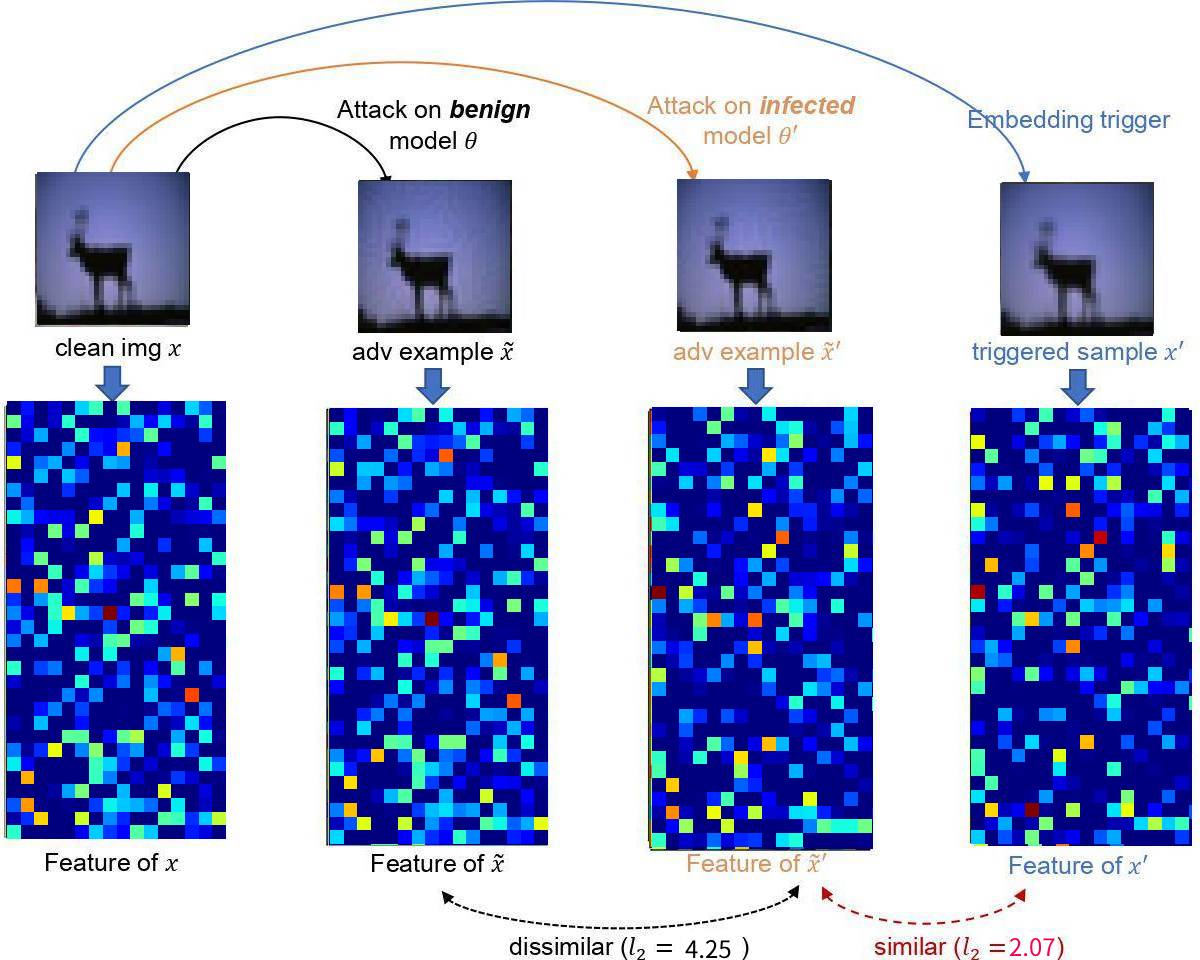}}
\caption{The two images are sampled from CIFAR-10, with size of $32 \times 32$, under WaNet Attack}
\label{featuremap_cifar10}
\end{figure*}

\subsubsection{Backdoor Attacks and Defense}

\noindent\textbf{Backdoor Attacks:}
We consider $5$ common backdoor attacks: 1) BadNets~\cite{gu2017badnets}, 2) Blend attack~\cite{chen2017blend} 3) Sinusoidal signal attack (SIG)~\cite{barni2019sig}, 4) Input-aware dynamic attack (DyAtt)~\cite{nguyen2020inputaware}, and 5) Warpping-based attack (WaNet)~\cite{nguyen2021wanet}. The first three belong to static backdoor attack, which has a fixed trigger (\emph{e.g.,} several pixels or an trigger image), while the latter two belong to the dynamic backdoor attack, which allows the trigger to adapt to the image content
(\emph{i.e.,} different images have different trigger patterns). The dynamic backdoor attacks are often much harder to defend against.

\noindent\textbf{Model Repairing-based Backdoor Defense:}
since our approach belongs to the category of \emph{model repairing} defense, we compare our PUD with $5$ existing model repairing methods: 1) the standard Fine-tuning~\cite{liu2017finetuning}, 2) Fine-pruning~\cite{liu2018finepurning}, 3) Neural Cleanse (NC)~\cite{wang2019neuralcleanse}, 4) Neural Attention Distillation (NAD)~\cite{li2020nad}, and 5) Adversarial Neuron Pruning (ANP)~\cite{ANP}. Although NC, NAD, and ANP have different defense strategies, they are all considered the state-of-the-art methods and have comparable performance.

\noindent\textbf{Data Filtering-based Backdoor Defense:}
there is another category of backdoor defense called \emph{data filtering}, which perform the defense \emph{during} (rather than \emph{after}) model training. They assume that they can access \emph{all} of the poisoned training images. They perform the defense by filtering out poisoned images in the training data and re-training a model from scratch, rather than erasing backdoors from infected models. 

%Note that the setting of data filtering methods is much strict than the model repairing category. 

With a simple pre-processing procedure, our approach can also work for the data filtering defense setting. Thus, we will also compare it to two existing data filtering methods, \emph{i.e.,} Spectral Signatures~\cite{tran2018spectral} and SPECTRE~\cite{hayase2021spectre}.

\subsubsection{Adversarial Attacks and Defenses}
To evaluate our approach against adversarial attacks, we consider the most popular adversarial attack, \emph{i.e.,} PGD-based attack~\cite{madry2018towards} in our experiments.% and the state-of-the-art AutoAttack~\cite{AutoAttack}. 

Regarding adversarial defense, Adversarial Training (AT)~\cite{madry2018towards} is the most representative defense method. Lately, it has been extended from different perspectives: changes in the attack (\emph{e.g.}, incorporating momentum~\cite{momentum}), loss function (\emph{e.g.}, logit pairing~\cite{pairing}) or model architecture (\emph{e.g.},
feature denoising~\cite{denoising}). Recently, another notable work is TRADES~\cite{TRADES}, which can balance the trade-off between standard and robust accuracy. 

In our experiments, we will compare our algorithm with these state-of-the-art adversarial defense to illustrate that our approach has competitive ability to defend against adversarial attacks.

% In addation, the poison ratio is set PR=$10\%$.
\subsubsection{Evaluation Metrics}
\noindent\textbf{Backdoor Robustness:}
we evaluate the performance of the backdoor defense using two metrics: the attack success rate (ASR), which is the ratio of triggered examples that are misclassified as the target label, and the model’s accuracy on clean samples (ACC). An ideal defense should result in large decreases in ASR with small penalties in ACC. 

\noindent\textbf{Adversarial Robustness:}
we also evaluate the performance of the adversarial defense with two metrics: Robust Accuracy (RACC), which is the accuracy on adversarial examples, and the model’s accuracy on clean samples (ACC). An ideal defense aims to improve RACC while keeping ACC as the same as that of a naturally trained model.

%\begin{figure*}[t]
%\centering
%\subfloat{
%\label{cm:1}
%\includegraphics[width=0.4\linewidth,height=0.3\textwidth]%{sup_figs/featuremap/feature_page-0003.jpg}}\hspace{10mm}
%\subfloat{
%\label{cm:2}
%\includegraphics[width=0.4\linewidth,height=0.3\textwidth]%{sup_figs/featuremap/feature_page_wa.jpg}}
%\caption{The two images are sampled from CIFAR-10, with size %of $32 \times 32$, under WaNet Attack}
%\label{featuremap_cifar10}
%\end{figure*}

%\begin{figure*}[t]
%\centering
%\subfloat{
%\includegraphics[width=0.3\linewidth,height=0.3\textwidth]{sup_figs/featuremap/feature_badnet-1.pdf}}\hspace{10mm}
%\subfloat{
%\includegraphics[width=0.3\linewidth,height=0.3\textwidth]{sup_figs/featuremap/feature_badnet-2.pdf}}
%\caption{The two images are sampled from CIFAR-10, with size of $32 \times 32$, under BadNet Attack.}
%\label{featuremap_badnet}
%\end{figure*}

%\begin{figure*}[t]
%\centering
%\subfloat{
%\label{cm:1}
%\includegraphics[width=0.4\linewidth,height=0.3\textwidth]{sup_figs/featuremap/feature_page-0001.jpg}}\hspace{10mm}
%\subfloat{
%\label{cm:2}
%\includegraphics[width=0.4\linewidth,height=0.3\textwidth]{sup_figs/featuremap/feature_page-0002.jpg}}
%\caption{Two images are sampled from ImageNet-1K, with size of $224 \times 224$, under Blend Attack.}
%\label{featuremap_imagenet}
%\end{figure*}

\subsection{Our Observation is Ubiquitous}
\subsubsection{Predicted labels v.s. Backdoor target-labels} Fig.\ref{cifar_hist} has illustrated one case (\emph{i.e.,} performing WaNet attack with the all-to-one setting on the CIFAR-10 dataset) about our observation that the adversarial examples are highly likely to be classified as the target-label. In this section, we illustrate that our observation is ubiquitous, \emph{i.e.,} it presents regardless of which attack method is employed (\emph{e.g.,} Blend, SIG, WaNet), which attack setting is used (\emph{e.g.,} All-to-one and All-to-all), and which dataset is used (\emph{e.g.,} CIFAR-10, GTSRB, ImgNet-1K).

As shown in Fig.\ref{cm}, we have similar observations: the predicted labels always align with the target-labels, as shown by the diagonal of the matrix. Due to the space limitations, for the ImgNet-1K dataset we only randomly select $10$ out of $1000$ classes. More results are shown in Appendix B.

In addition, we also examine the adversarial images generated with different adversarial attack settings. Except for the $l_2$ norm attack, we also evaluate the $l_{\inf}$ attack (\emph{i.e.,} set $p=l_{\inf}$ in Eq.\eqref{eq:adv2}). We also use different attack methods (such as PGD attack, C\&W attack) to generate adversarial examples. Finally, we find that our observation is ubiquitous, \emph{i.e.,} it presents regardless of which adversarial attack setting is used, and which adversarial attack method is used.

%\begin{table}[htpb]
%\scriptsize
%\centering
%\caption{\centering Feature Similarity on CIFAR-10.}
%\resizebox{0.4\textwidth}{!}{
  %\begin{tabular}{c | c | c | c |c | c}
  %\toprule
    %& Badnet &  SIG & Blend & DyAtt & WaNet \\
    %\midrule
    %$D_{\text{benign}}$ & 102.58 & 135.91 & 124.22 & 40.42 & 48.13 %\\
    %$D_{\text{infected}}$ & 85.11 &  78.18  & 75.09 & 28.66 & 15.85 \\
%\bottomrule
  %\end{tabular}
  %}
%\label{feature_similarity} 
%\vspace{-0.5em}
%\end{table}

%\begin{table}[htpb]
%\scriptsize
%\centering
%\caption{\centering Feature Similarity on GTSRB dataset.}
%\resizebox{0.4\textwidth}{!}{
  %\begin{tabular}{c | c | c | c |c | c}
  %\toprule
    %& Badnet &  SIG & Blend & DyAtt & WaNet \\
    %\midrule
    %$D_{\text{benign}}$ &162.07   &166.77   &150.87   &174.19   %&169.31   \\
    %$D_{\text{infected}}$ & 119.79  &50.46     &66.32  &160.81    %&102.72   \\
%\bottomrule
  %\end{tabular}
  %}
%\label{feature_similarity2} 
%\vspace{-0.0em}
%\end{table}

\subsubsection{Comparisons of Feature Similarity} 
Fig.\ref{backdoor_feat} is just an example of how the features of an adversarial image $\widetilde{\bm{x}}'$ (related to an infected model) are very similar to the features of the triggered image $\bm{x^t}$, rather than the original adversarial image $\widetilde{\bm{x}}$ (related to a benign model). More examples are given in this section (as shown in Fig.\ref{featuremap_cifar10}) and in Appendix C.

\begin{table}[htpb]
\scriptsize
\centering
\caption{\centering Feature Similarity on CIFAR-10.}
\resizebox{0.4\textwidth}{!}{
  \begin{tabular}{c | c | c | c |c | c}
  \toprule
    & Badnet &  SIG & Blend & DyAtt & WaNet \\
    \midrule
    $D_{\text{benign}}$ & 2.09  &1.86  & 1.60 &  1.64 & 2.64\\
    $D_{\text{trigger}}$ & 0.97 & 0.93   & 0.82 & 0.75 &  0.90\\
\bottomrule
  \end{tabular}
  }
\label{feature_similarity} 
\vspace{-1.0em}
\end{table}

\begin{table}[htpb]
\scriptsize
\centering
\caption{\centering Feature Similarity on GTSRB dataset.}
\resizebox{0.4\textwidth}{!}{
  \begin{tabular}{c | c | c | c |c | c}
  \toprule
    & Badnet &  SIG & Blend & DyAtt & WaNet \\
    \midrule
    $D_{\text{benign}}$ &1.18   &1.41   &1.10   &1.45   &1.84   \\
    $D_{\text{trigger}}$ & 0.87  &0.33     &0.22  &1.06    &1.10   \\
\bottomrule
  \end{tabular}
  }
\label{feature_similarity2} 
\vspace{1em}
\end{table}

All of the previous figures only illustrate feature similarity qualitatively, we further make quantitative comparisons. Specifically, we randomly sample $10,000$ images from CIFAR-10, and compute the $l_2$ distances between the features of $\widetilde{\bm{x}}'$ and $\bm{x^t}$, 
\begin{equation}
    D_{\text{trigger}}=||f(\widetilde{\bm{x}}'),f(\bm{x^t})||_2
    \label{eq:infected},
\end{equation}
Meanwhile, we also calculate the $l_2$ distances between the features of $\widetilde{\bm{x}}'$ and $\widetilde{\bm{x}}$, 
\begin{equation}
    D_{\text{benign}}=||f(\widetilde{\bm{x}}'),f(\widetilde{\bm{x}})||_2
    \label{eq:benign},
\end{equation}
where the image features $f()$ is the output of the last convolution layer (just before the fully-connected layer). This evaluation includes BadNet, SIG, Blend, DynAtt, and WaNet.

By comparing $D_{\text{trigger}}$ and $D_{\text{benign}}$, we can quantitatively find that: after planting a backdoor in a model, the feature $f(\widetilde{\bm{x}}')$ is much more similar to the feature $f(\bm{x^t})$ than the original feature $f(\widetilde{\bm{x}})$, as shown in Table \ref{feature_similarity} and Table \ref{feature_similarity2}.

\begin{figure}[htpb]
%\vspace{-0.5em}
\centering
\includegraphics[width=1.0\linewidth]{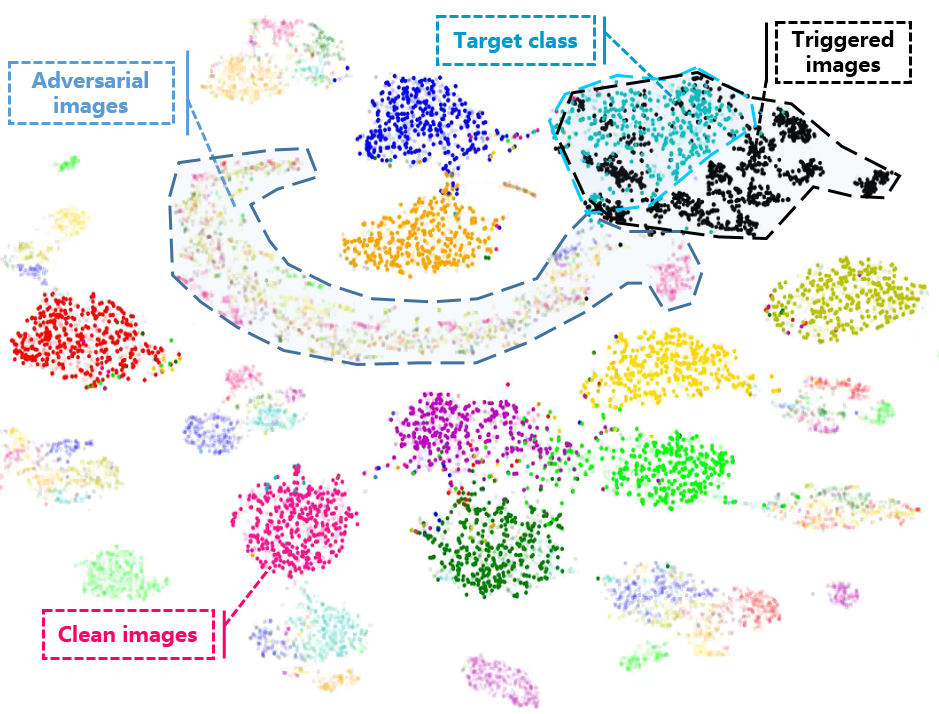}
\caption{The visualization of image features from CIFAR-10. The original images are in dark color (\emph{e.g.,} dark-red), while the corresponding adversarial examples are in light color (\emph{e.g.,} light-red). In this case, the target class is shown in cyan. The triggered images are shown in black, which is close to target class. Obviously, lots of adversarial examples \emph{lie on a `belt' which is close to the triggered images}.}
\label{backdoor_tsne}
\vspace{-0.5em}
\end{figure}

We also visualize these high-dimensional features in a 2D space using t-SNE~\cite{tSNE}. Fig.\ref{backdoor_tsne} shows the features of (a) the original clean images in dark colors; (b) their corresponding adversarial images in light colors; and (c) the triggered images (in black). In this case, the backdoor target-label is shown in cyan.

From Fig.\ref{backdoor_tsne}, it is obvious that most of the adversarial images, triggered images, and target-label images lie on the same data manifold. This also justifies that the adversarial images are very similar to the triggered images.

\begin{table*}[tpb]
\scriptsize
 \centering
 \caption{Comparison with SoTA defense methods on CIFAR-10 (all-to-one setting). All 5 methods and our AFT \emph{use} the extra \emph{clean} dataset, while the PBE and PUD \emph{do not use} such extra \emph{clean} dataset. Instead, both PBE and PUD just use a \emph{small poisoned} extra dataset. For each backdoor attack, the \textbf{best ASR} are highlighted. `F' means fail to defense.}
 \resizebox{1.02\textwidth}{!}{
 \setlength{\tabcolsep}{0.3em}%
  \begin{tabular}{c|c c|c c|c c|c c|c c|c c|c c|c c|c c|c c}
  \hline
  \multirow{0}* & 
  \multicolumn{2}{c|}{Before} & \multicolumn{2}{c|}{Fine-tuning} & 
  \multicolumn{2}{c|}{Fine-pruning} & \multicolumn{2}{c|}{NAD} & 
  \multicolumn{2}{c|}{Neural Cleanse} &
  \multicolumn{2}{c|}{ANP} & 
  \multicolumn{2}{c|}{\myblue{I-BAU}} &
  \multicolumn{2}{c|}{\textbf{AFT}} &   
  \multicolumn{2}{c|}{\textbf{PBE}} & 
  \multicolumn{2}{c}{\textbf{PUD}} \\
  \cline{2-21}
  & ACC & ASR
  & ACC$\uparrow$ & ASR$\downarrow$
  & ACC$\uparrow$ & ASR$\downarrow$
  & ACC$\uparrow$ & ASR$\downarrow$
  & ACC$\uparrow$ & ASR$\downarrow$
  & ACC$\uparrow$ & ASR$\downarrow$
  & \myblue{ACC$\uparrow$} & \myblue{ASR$\downarrow$}
   & ACC$\uparrow$ & ASR$\downarrow$
   & ACC$\uparrow$ & ASR$\downarrow$   
  & ACC$\uparrow$ & ASR$\downarrow$\\
  \hline
    Badnet & 94.67 & 100.00 & 85.82 & 6.53 & 89.80 & 70.66 & 88.09 & 2.17 & 93.73 & \textbf{0.83} & 93.39 & 1.66 &\myblue{88.61} &\myblue{9.08} &94.20 &1.09 &87.91 &3.41 &
    93.34 &1.44\\
    Blend & 94.63 & 100.00 & 87.53 & 11.31 & 89.30 & 65.86 & 90.13& 1.60 & 93.28 & 0.63 & 92.03 & 1.81 &\myblue{90.14} &\myblue{16.59}&93.98 &0.93 &90.88  &1.15 &92.01 &\textbf{0.55}\\ 
    SIG & 94.81 & 98.96 & 87.34 & 4.14 & 88.93 & 85.69 & 90.26 & 4.59 & 92.23 & 1.79 & 92.48 & 1.27  &\myblue{89.15} &\myblue{4.45}& 93.35 & 1.39 &89.35 &3.53  &92.48 &\textbf{0.31} \\ 
    DyAtt & 94.65 & 99.24 & 94.00 & 8.77 & 89.91 & 98.97 & 94.23 & 4.59 & 94.65 & F & 93.42 & 1.36 &\myblue{87.09} &\myblue{2.04} & 93.01 & 1.12 &92.33 &0.57 &92.32 &\textbf{0.53}\\ 
    WaNet & 94.15 & 99.50 & 93.42 & 12.80 & 89.86 & 99.36 & 94.02 & 8.37 & 94.15 & F & 93.36 & 0.62 &\myblue{90.06} &\myblue{7.17}&94.32 &\textbf{0.46} &93.84 &0.83 &94.07 &0.76 \\ 
  \hline
  \end{tabular}
}
\label{tb1} 
\end{table*}

\begin{table*}[tpb]
\scriptsize
 \centering
 \caption{Comparison with SoTA defense methods on GTSRB dataset (all-to-one setting).} 
 \resizebox{1.02\textwidth}{!}{
 \setlength{\tabcolsep}{0.3em}%
  \begin{tabular}{c|c c|c c|c c|c c|c c|c c|c c|c c|c c|c c}
  \hline
  \multirow{0}* & 
  \multicolumn{2}{c|}{Before} & \multicolumn{2}{c|}{Fine-tuning} & 
  \multicolumn{2}{c|}{Fine-pruning} & \multicolumn{2}{c|}{NAD} & 
  \multicolumn{2}{c|}{Neural Cleanse} & 
  \multicolumn{2}{c|}{ANP} & 
  \multicolumn{2}{c|}{\myblue{I-BAU}} &
  \multicolumn{2}{c|}{\textbf{AFT}} & 
  \multicolumn{2}{c|}{\textbf{PBE}} &   
  \multicolumn{2}{c}{\textbf{PUD}} \\
  \cline{2-21}
  & ACC & ASR
  & ACC$\uparrow$ & ASR$\downarrow$
  & ACC$\uparrow$ & ASR$\downarrow$
  & ACC$\uparrow$ & ASR$\downarrow$
  & ACC$\uparrow$ & ASR$\downarrow$
  & ACC$\uparrow$ & ASR$\downarrow$
  & \myblue{ACC$\uparrow$} & \myblue{ASR$\downarrow$}
  & ACC$\uparrow$ & ASR$\downarrow$
  & ACC$\uparrow$ & ASR$\downarrow$  
  & ACC$\uparrow$ & ASR$\downarrow$ \\
  \hline
    Badnet & 99.02 & 100.00 & 95.01 & 8.32 & 89.60 & 75.03 & 94.22 & 2.06 & 96.78 & \textbf{0.12} & 95.13 & 1.35 &\myblue{99.32} &\myblue{1.17}& 94.43 & 0.47  &95.24  &2.91 &97.54 &0.92 \\
    Blend & 99.39 & 99.92 & 90.68 & 40.12 & 88.21 & 90.53 & 92.61 & 8.56 & 96.48 & 5.81& 94.02 & 2.68 &\myblue{97.38} &\myblue{4.42}& 94.57 & 1.72 &91.27  &2.39  &98.78 &\textbf{0.45} \\ 
    SIG & 98.56 & 95.81 & 91.63 & 36.30 & 89.53 & 93.26 & 92.94 & 6.90 & 93.40 & 1.32 & 93.32 & 3.65 &\myblue{97.89} &\myblue{4.65}& 94.05 & 1.78 &89.81  &2.89  &97.04 &\textbf{0.37}\\ 
    DyAtt & 99.27 & 99.84 & 97.10 & 16.33 & 89.15 & 97.21 & 98.17 & 3.80 & 99.27 & F &95.88 & 1.68 &\myblue{97.53} &\myblue{0.06} & 96.68 & 0.99 &95.99  &1.63  &98.70 &\textbf{0.21} \\ 
    WaNet & 98.97 & 98.78 & 96.70 & 4.20 & 87.49 & 98.79 & 97.07 & 2.20 & 98.97 & F & 96.47 & 0.94 &\myblue{97.40} &\myblue{0.49}& 96.56 & 0.47 &95.23 &0.64  &98.76 &\textbf{0.31}\\ 
%   \hline
\bottomrule
  \end{tabular}
 }
\label{tb2} 
\end{table*}

\begin{table*}[tpb]
\scriptsize
 \centering
 \caption{Comparison with SoTA defense methods on sub-ImgNet1K dataset (all-to-one setting).} 
 \resizebox{1.02\textwidth}{!}{
 \setlength{\tabcolsep}{0.3em}%
  \begin{tabular}{c|c c|c c|c c|c c|c c|c c|c c|c c|c c|c c}
  \hline
  \multirow{0}* & 
  \multicolumn{2}{c|}{Before} & \multicolumn{2}{c|}{Fine-tuning} & 
  \multicolumn{2}{c|}{Fine-pruning} & \multicolumn{2}{c|}{NAD} & 
  \multicolumn{2}{c|}{Neural Cleanse} & 
  \multicolumn{2}{c|}{ANP} & 
  \multicolumn{2}{c|}{\myblue{I-BAU}} &
  \multicolumn{2}{c|}{\textbf{AFT}} & 
  \multicolumn{2}{c|}{\textbf{PBE}} &   
  \multicolumn{2}{c}{\textbf{PUD}} \\
  \cline{2-19}
  & ACC & ASR
  & ACC$\uparrow$ & ASR$\downarrow$
  & ACC$\uparrow$ & ASR$\downarrow$
  & ACC$\uparrow$ & ASR$\downarrow$
  & ACC$\uparrow$ & ASR$\downarrow$
  & ACC$\uparrow$ & ASR$\downarrow$
  & \myblue{ACC$\uparrow$} & \myblue{ASR$\downarrow$}
  & ACC$\uparrow$ & ASR$\downarrow$
  & ACC$\uparrow$ & ASR$\downarrow$  
  & ACC$\uparrow$ & ASR$\downarrow$ \\
  \hline
    Badnet &97.37  &99.17  &96.86  &95.84  & 89.68 & 98.1  &94.72  &51.04  &81.63  &\textbf{0.75}  &92.27 &2.96  &\myblue{93.60} &\myblue{32.0} &92.78  &1.69  &87.36  &1.68 &92.45  &1.52\\
    Blend &97.01  &97.90  &95.28 &7.49 &  93.12 & 88.40  &94.80  &5.53  &80.22 &2.82 &94.24  &0.73 &\myblue{88.40} &\myblue{55.77} &94.75  &0.71  &91.94 &1.03 &95.13  &\textbf{0.65}  \\ 
    SIG &97.52  &99.91  &96.33  &63.56  & 87.71 & 91.63  &94.36   &22.62  & 85.20  & 1.13  &94.50  &3.86  &\myblue{90.80} &\myblue{6.22} &95.23 &0.70 &94.11
      &1.98 &94.49   &\textbf{0.42}\\ 
    DyAtt &94.01  &88.53  &93.07  &6.50  & 89.57 & 86.32  &92.99  &0.14  &94.01  & F  &88.63 &2.74  &\myblue{90.82} &\myblue{29.01} &92.56  &0.91   &91.69 &1.55   &92.96 &\textbf{0.73}\\ 
    WaNet &97.35  &86.06  &95.33  &1.29  & 91.29  & 85.94  &94.49  &0.40  &97.35  & F  &89.25  &3.87  &\myblue{90.34} &\myblue{1.18}&95.08  &0.48  &95.21 &0.96 &96.28 &\textbf{0.31}\\ 
%   \hline
\bottomrule
  \end{tabular}
 }
\label{tb3} 
%\vspace{1.5em}
\end{table*}

\begin{table*}[tpb]
\scriptsize
 \centering
 \caption{Comparison with SoTA defense methods on CIFAR-10 (all-to-all setting) .}
 \resizebox{1.02\textwidth}{!}{
 \setlength{\tabcolsep}{0.3em}%
  \begin{tabular}{c|c c|c c|c c|c c|c c|c c|c c|c c|c c|c c}
  \hline
  \multirow{0}* & 
  \multicolumn{2}{c|}{Before} & \multicolumn{2}{c|}{Fine-tuning} & 
  \multicolumn{2}{c|}{Fine-pruning} & \multicolumn{2}{c|}{NAD} & 
  \multicolumn{2}{c|}{Neural Cleanse} & \multicolumn{2}{c|}{ANP} & 
  \multicolumn{2}{c|}{\myblue{I-BAU}} &
   \multicolumn{2}{c|}{\textbf{AFT}} & 
   \multicolumn{2}{c|}{\textbf{PBE}} &    
  \multicolumn{2}{c}{\textbf{PUD}} \\
  \cline{2-21}
  & ACC & ASR
  & ACC$\uparrow$ & ASR$\downarrow$
  & ACC$\uparrow$ & ASR$\downarrow$
  & ACC$\uparrow$ & ASR$\downarrow$
  & ACC$\uparrow$ & ASR$\downarrow$
  & ACC$\uparrow$ & ASR$\downarrow$
  & \myblue{ACC$\uparrow$} & \myblue{ASR$\downarrow$}
  & ACC$\uparrow$ & ASR$\downarrow$
  & ACC$\uparrow$ & ASR$\downarrow$  
  & ACC$\uparrow$ & ASR$\downarrow$ \\
  \hline
    Badnet & 94.63 & 94.41 & 85.54 & 3.68 & 88.20 & 66.13 & 89.63 & 1.01 & 94.63 & 94.41 & 92.01 & 0.69 &\myblue{87.16} &\myblue{8.85} &  93.84 & \textbf{0.62} &91.37  &0.91  &91.62  &0.64  \\
    Blend & 94.89 & 87.94 & 86.60 & 5.36 & 87.96 & 74.15 & 89.91 & 2.38 & 94.89 & 87.94 & 93.12 & 1.24 &\myblue{86.98} &\myblue{2.92} & 93.65 & 0.68 &91.16  &2.17  &92.46  &\textbf{0.60}  \\ 
    SIG & 94.66 & 84.34 & 87.98 & 2.83 & 88.99 & 69.52 & 91.53 & 1.36 & 94.66 & 84.34 & 93.60 & 0.87 &\myblue{86.72} &\myblue{2.58} &93.52 & \textbf{1.01} &89.36 &2.02 
    &92.06 &1.32\\
    DyAtt & 94.40 & 92.72 & 92.05 & 4.46 & 89.62 & 90.33 & 92.71 & 1.39  & 94.40 & F & 92.86 & 1.09 &\myblue{87.19} &\myblue{0.87}& 93.28 & 0.75 &89.19 &4.56 &93.49 &\textbf{0.61} \\ 
    WaNet & 94.49 & 93.47 & 93.37 & 7.81 & 89.02 & 92.53 & 93.68 & 3.05 & 94.49 & F & 93.21 & 0.99 &\myblue{87.04} &\myblue{1.33}& 93.45 & 0.80 &93.18  &0.88  &93.23  &\textbf{0.75} \\
%   \hline
\bottomrule
  \end{tabular}
 }
\label{all2all}
\end{table*}

\iffalse

\begin{table*}[tpb]
\scriptsize
 \centering
 \caption{Comparison with SoTA defense methods on GTSRB dataset (\textbf{all-to-all setting}).} 
 \resizebox{1.02\textwidth}{!}{
 \setlength{\tabcolsep}{0.3em}%
  \begin{tabular}{c|c c|c c|c c|c c|c c|c c|c c|c c|c c}
  \hline
  \multirow{0}* & 
  \multicolumn{2}{c|}{Before} & \multicolumn{2}{c|}{Fine-tuning} & 
  \multicolumn{2}{c|}{Fine-pruning} & \multicolumn{2}{c|}{NAD} & 
  \multicolumn{2}{c|}{Neural Cleanse} & 
  \multicolumn{2}{c|}{ANP} & 
  \multicolumn{2}{c|}{\textbf{AFT}} & 
  \multicolumn{2}{c|}{\textbf{PBE}} &   
  \multicolumn{2}{c}{\textbf{PUD}} \\
  \cline{2-19}
  & ACC & ASR
  & ACC$\uparrow$ & ASR$\downarrow$
  & ACC$\uparrow$ & ASR$\downarrow$
  & ACC$\uparrow$ & ASR$\downarrow$
  & ACC$\uparrow$ & ASR$\downarrow$
  & ACC$\uparrow$ & ASR$\downarrow$
  & ACC$\uparrow$ & ASR$\downarrow$
  & ACC$\uparrow$ & ASR$\downarrow$  
  & ACC$\uparrow$ & ASR$\downarrow$ \\
  \hline
    Badnet &98.38  &97.76  &  &  &  &  &  &  &  &  &  &  &  &  &\red{92.69}  &\red{6.8} &\red{95.48}  &\red{0.21}\\
    Blend &98.93  &98.75  & & &  &  &  &  &  & &  & &  &  &\red{98.21} &\red{0.89} &\red{98.32}  &\red{0.77}  \\ 
    SIG &  &  &  &  &  &  &  &  &  &  &  &  & &  &  &  \\ 
    DynAtt &  &  &  &  &  &  &  &  &  &  & &  &  &  &  &   \\ 
    WaNet &99.09  &93.58  &  &  &  &  &  &  &  &  &  &  &  &  &\red{99.03}  &\red{0.12} &\red{99.18} &\red{0.05}\\ 
%   \hline
\bottomrule
  \end{tabular}
 }
\label{all2all2} 
%\vspace{1.5em}
\end{table*}

\fi

\subsection{Comparison to Model Repairing-based Defenses}
We compare our PUD with $5$ state-of-the-art model repairing methods: Fine-tuning~\cite{liu2017finetuning}, Fine-pruning~\cite{liu2018finepurning}, NC~\cite{wang2019neuralcleanse}, NAD~\cite{li2020nad}, and ANP~\cite{ANP}. All these $5$ defense methods need to have an extra \emph{clean} dataset. In contrast, our approach can erase backdoors either with or without an extra clean dataset. We will discuss each of these settings, respectively. 

\subsubsection{With an extra clean dataset}~\label{with_extra}
In this case, our PUD algorithm is simplified as the AFT algorithm (refer to Section \ref{AFT}). Note that when an extra clean dataset is given, both PUD and PBE are simplified as the same algorithm, \emph{i.e.,} AFT algorithm.

Regarding the extra clean dataset, we follow the common experimental settings, which is randomly sampled from the \emph{clean} training data, taking about $5\%$ of it. 
The comparisons between the $5$ previous methods and our AFT are fair under this setting, since they all use the same extra clean dataset. 

The comparisons on CIFAR-10, GTSRB, and sub-ImgNet1K datasets are shown in Table.\ref{tb1}, \ref{tb2} and \ref{tb3}. Our approach  (\emph{i.e.,} AFT version) can remarkably reduce ASR while maintaining ACC. It significantly outperforms all existing defenses against all attacks except the Neural Cleanse (NC) defense under Badnet attack. 

Regarding the NC defense, it has quite good performance against Badnet attack, but cannot defend against the content-aware attacks (\emph{e.g.,} DyAtt, WaNet), as shown by `F'(Fail to defense) in Table.\ref{tb1}, \ref{tb2} and \ref{tb3}. This is because NC is a trigger synthesis-based method, so it needs to recover a trigger. However, content-aware attacks make triggers adapt to the image content (instead of using a fixed trigger), and thus are able to circumvent the NC defense.

\subsubsection{Without an extra clean dataset}
Both the PBE and PUD algorithms can erase backdoors without an extra \emph{clean} dataset. We simply assume to have an extra dataset and allow it to contain \emph{poisoned} images. In practice, we can obtain such an extra dataset by sampling from the poisoned training data. Specifically, in our experiments, we randomly sample $10\%$ of all training data to form the extra dataset. 
%The PBE assumes to have the \emph{whole} poisoned training data, while a \emph{small} part of it is enough for PUD. For fair comparison, the PBE in this experiments uses the same extra dataset as the PUD. 

%If an extra clean dataset is unavailable, all the $5$ defense methods \textbf{cannot} work, but our approach (\emph{i.e.,} both PBE and PUD versions) could still achieve excellent defensive performance. Notice that, PBE needs to use \emph{all} poisoned training data while PUD just use \emph{$10\%$} of poisoned training data, our PUD still outperforms PBE due to the three advancements about data and model purification. The ablation study of the three advancements will be described in the next section.

From Table.\ref{tb1} and \ref{tb2}, we can see that our PUD outperforms PBE for all types of backdoor attacks. In particular, for Badnet, Blend, and SIG attacks, PUD significantly outperforms PBE, \emph{e.g.,} reducing ASR from $3.41\%$ to $1.44\%$ for Badnet attack, from $1.15\%$ to $0.55\%$ for Blend attack, from $3.53\%$ to $0.31\%$ for SIG attack. The advantage of PUD over PBE is due to the proposed data and model purification mechanisms.

Besides, it is worth noting that even without using clean extra dataset, our PUD outperforms AFT for Blend, DyAtt attacks. It means that with proper data purification mechanism, $10\%$ poisoned images can help to achieve better backdoor erasing performance than $5\%$ clean images.

Similar to Sec~\ref{with_extra}, our PUD is slightly weaker than NC defense against Badnet attacks. This is because Badnet can efficiently backdoor a model using only a few poisoning images, while both PBE and PUD cannot $100\%$ filter out all the poisoning images in the extra dataset. 

\subsubsection{Large image resolution}
The images in CIFAR-10 and GTSRB dataset are $32\times 32$ in size. To evaluate our approach on high resolution images, we perform experiments on the ImageNet dataset ($224\times 224$). We randomly select 10 classes out of 1,000 classes for evaluation. Resnet-18 is used as the classification model. From Table.\ref{tb3}, we can see that our PUD still outperforms other methods for large image resolution settings.

\subsubsection{Results for all-to-all setting} 
There are two attack settings for backdoor attacks: \emph{All-to-one} setting (the target-labels for all examples are set to the same label) and \emph{All-to-all} setting (the target-labels for different classes could be set differently). All previous experiments are done for the all-to-one setting, and this section will discuss the all-to-all setting.

Following the previous methods~\cite{nguyen2021wanet}, we set the target-label to $y+1$ for the all-to-all setting. From Table.\ref{all2all}, it is obvious that our approach is very effective in this attack setting. Note that Neural Cleanse has poor defensive performance for the all-to-all attack setting.

%\noindent\textbf{Results for large-scale dataset:} To evaluate our approach on a large-scale dataset, we conduct experiments on the Tiny-ImageNet dataset~\cite{tinyimagenet}. The original Tiny-ImageNet contains $100,000$ images of 200 classes downsized to 64×64 colored images. In order to reload the pre-trained weights associated to the standard ImageNet dataset~\cite{deng2009imagenet}, we resize those images back to 224×224, and adopt Resnet-18 as the baseline classification model. From Table.\ref{tinyimagenet}, We can see that our approach is still effective on a large-scale dataset.

\subsection{Progressive Learning}
Progressive learning is the special property of our approach, we will discuss the progressive data purification and the progressive model purification, respectively.
\subsubsection{Prediction-based Data Purification}
In our approach, we formulate the data purification as an image ranking problem. Thus, we can evaluate its performance using the Precision-Recall curve and the Average Precision (AP) score. 

Specifically, for each image $\bm{x}_i \in D_{ext}$, we rank them in descending order according to their prediction consistency scores $C_{\theta', \eta^t}(\bm{x})$ as Eq.\eqref{eq:idclean}. The higher an image is ranked, the more likely it is regarded as an clean image by our approach. Therefore, given the ground-truth of $\bm{x}_i$ (\emph{i.e.,} clean or poisoned), we can draw the PR curve and evaluate our purification performance according to the ranking results.

\begin{figure}[h]
\vspace{-1em}
\centering
\includegraphics[width=0.9\linewidth]{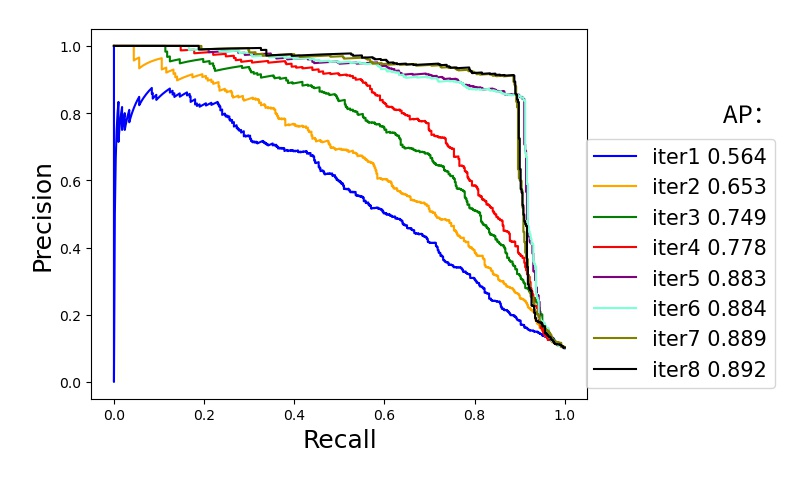}
\vspace{-1em}
\caption{The Progress of prediction-based data purification w.r.t. the increase of iterations for blend attack on CIFAR-10.}
\label{fig:pr}
\end{figure}

\begin{figure*}[t]
\centering
\subfloat[ASR]{
\label{mp:1}
\includegraphics[width=0.3\linewidth]{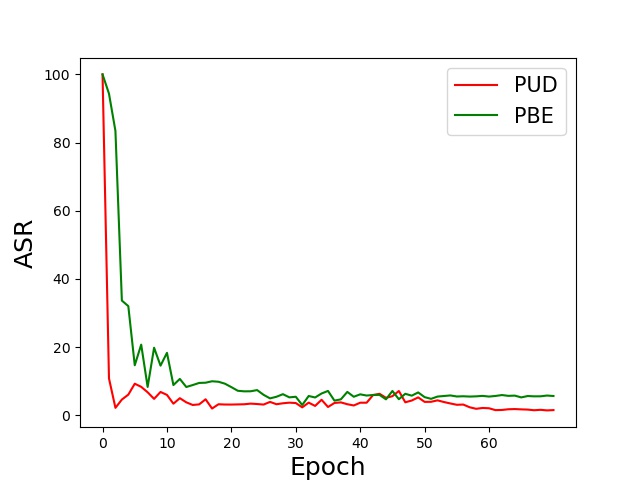}}\hspace{1mm}
\subfloat[ACC]{
\label{mp:2}
\includegraphics[width=0.3\linewidth]{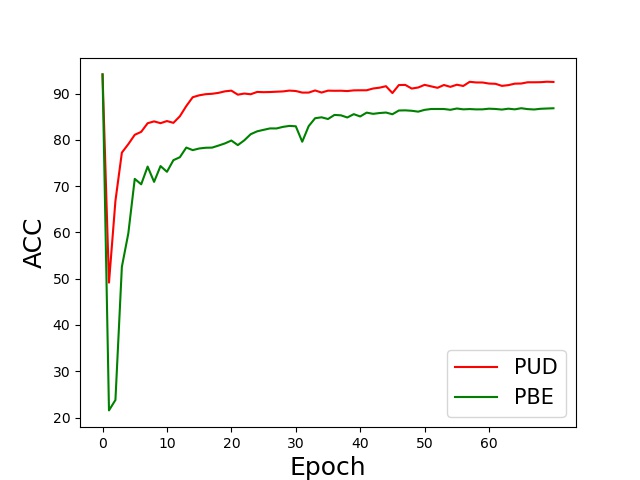}}\hspace{1mm}
\subfloat[RACC]{
\label{mp:3}
\includegraphics[width=0.3\linewidth]{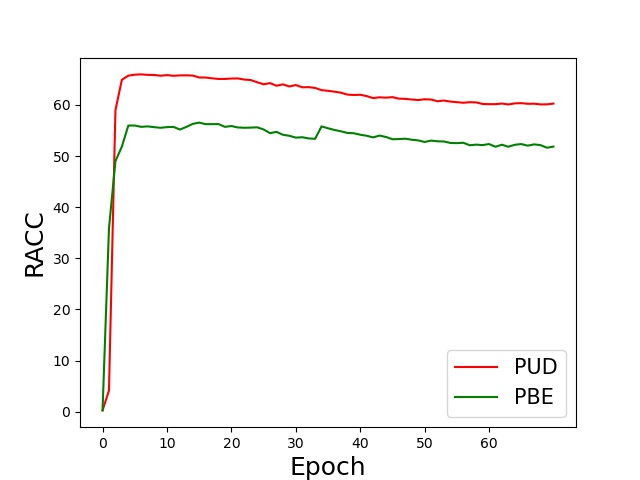}}\hspace{10mm}
\caption{The Progress of model purification (in terms of ASR, ACC, and RACC) w.r.t. the increase of iterations.}
\label{fig:PUD_model}
\end{figure*}

Due to the progressive nature of our approach, the performance of our data purification is gradually improved. Fig.\ref{fig:pr} is an example of such progress. As the iterations increases, the PR curve as well as the AP scores become better and better. This indicates that the quality of clean image identification is progressively improved.

\subsubsection{Progress of Data Purification}
\begin{figure}[htpb]
\vspace{-1em}
\centering
\includegraphics[width=0.7\linewidth]{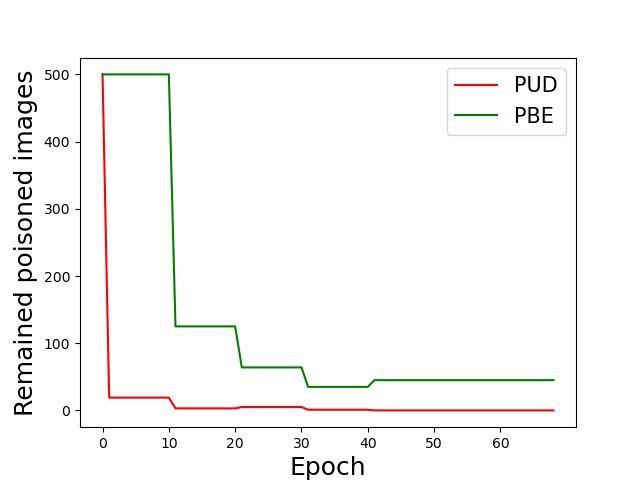}
\caption{The Progress of final data purification w.r.t. the increase of iterations.}
\label{fig:PUD_data}
\end{figure}

Our PUD combines the prediction-based strategy with the SPECTRE-based strategy for final data purification. With the combination of two purification strategies, the quality of the purified extra dataset is progressively improved.
From Fig.\ref{fig:PUD_data}, we observe that with an increase in iterations, more poisoned images in the extra dataset are filtered out. 

Particularly, the prediction-based strategy is complementary to the SPECTRE-based strategy, and their combination can significantly improve the performance of data purification. 
From Table.\ref{tb:dp}, we can see that our PUD is better than both the prediction-based strategy (\emph{i.e.,} PBE) and the SPECTRE alone.

\begin{table}[htpb]
\vspace{-1em}
\centering
\caption{\centering Number of filtered poisoned images on CIFAR-10. There are $500$ poisoned images in total.} 
\label{filter}
 \resizebox{0.5\textwidth}{!}{
\begin{tabular}{cccc}
    \toprule
     & Prediction-based (\emph{i.e.,} PBE) & SPECTRE & PUD \\
    \midrule
    Badnet & 455/500 & 488/500  & \textbf{499}/500  \\
    Blend  & 483/500 & 492/500  & \textbf{500}/500  \\
    SIG    & 473/500 & 489/500  & \textbf{500}/500  \\
    %Blind    & 484/500 & 489/500  & \textbf{500}/500  \\
    %AdvDoor    & /500 & 330/500  & \textbf/500  \\
    \bottomrule
\end{tabular}
}
\label{tb:dp}
\vspace{-1em}
\end{table}

%\noindent\textbf{Ablation study of model purification:} comparing to PBE, one advantage of our PUD is the teacher-student mechanism, which can better purify model. From Table.\ref{tb1}, \ref{tb2}, we can see that our PUD outperforms 

\subsubsection{Progress of Model Purification}
In our approach, the quality of the purified model is progressively improved with the increase of iterations. 

From Fig.\ref{mp:1} and Fig.\ref{mp:2}, we can see that with the increase of iterations, the ASR decreases while ACC increases gradually, indicating that the purified model is getting better and better on both benign and poisoned images. Particularly, the backdoor can be quickly erased at beginning iterations (ASR decreases quickly), while the following iterations mainly help to improve the performance on benign images (ACC increases gradually). Obviously, PUD outperforms PBE in terms of backdoor robustness.

In addition, as shown in Fig.\ref{mp:3}, the RACC also increases with the increase of iterations, indicating that the model's adversarial robustness is gradually improved. Moreover, PUD significantly outperforms PBE in terms of adversarial robustness. The most advantage of PUD over is PBE is that it can simultaneously improve the model's backdoor robustness as well as the model's adversarial robustness.

\subsection{Ablation Study}
The previous results show that PUD has better robustness against both backdoor and adversarial attacks than PBE. This is due to the following three technical advances: (1) A mean-teacher mechanism is used to enhance the robustness against both backdoor and adversarial attacks; (2) SPECTRE is used to complement prediction-based data purification; (3) backdoor unlearning scheme is used to utilize poisoned images for further backdoor erasing.

We perform an ablation study to illustrate the individual contribution from each technical point. From Table.\ref{tb:ab}, we can see that the mean-teacher mechanism and SPECTRE-based strategy can significantly improve the model's adversarial robustness (\emph{i.e.,} improving RACC). \myblue{It is because we utilize SPECTRE to filter out high-confidence poisoned images, resulting in more images remaining in $D_{ext}^t$.
Thus, the fine-tuning step can leverage these images to enhance adversarial robustness, contributing to a high RACC. However, the presence of poisoned images within the remaining data is detrimental to backdoor erasing, resulting in a worse ASR.}

On the other hand, the backdoor unlearning can significantly boost the model's backdoor robustness (\emph{i.e.,} reducing ASR). Notably, our PUD combines the three technical advances so that it can simultaneously improve the two types of model robustness.

\begin{table}[htpb]
\centering
\caption{\centering Ablation study for Blend attack on CIFAR-10.} 
\label{filter}
 \resizebox{0.5\textwidth}{!}{
\begin{tabular}{ccccc}
    \toprule
     & ACC & ASR & RACC &\myblue{Overhead}\\
    \midrule
    Before &94.67  &100.0  & 0.24 & \\
    PBE  &90.88  &1.15  &50.64 & \myblue{346}\\
    PBE + mean-teacher    &90.36  &0.84  &56.74 & \myblue{419}\\
    PBE + SPECTRE &92.36  &1.02  &\textbf{61.13} & \myblue{457}\\
    PBE + Unlearning &90.61  &\textbf{0.48}  &50.67 & \myblue{401}\\    
    PUD &92.01  &0.55 & 59.35 & \myblue{664}\\
    \bottomrule
\end{tabular}
}
\vspace{-2em}
\label{tb:ab}
\end{table}

\subsection{Comparison to Data Filtering-based Defenses}\label{sec:filtering}
Since our approach has the ability of data purification, it can also work with a data filtering defensive setting. Specifically, we adopt a pre-pressing procedure (\emph{i.e.,} Algorithm 2) before running PUD. 

For comparison, two common data filtering defense are considered in this paper, \emph{i.e.,} Spectral Signatures~\cite{tran2018spectral} and SPECTRE~\cite{hayase2021spectre}. According to their settings, $500$ training images are randomly selected to be poisoned. 
Then, the two defenses try to filter out the poisoned images in training data and re-train a model from scratch.

For performance evaluation, we first directly compare the results of data filtering, \emph{i.e.,} how many poisoned training images are correctly filtered out. And then, we compare the final defense performance in terms of ASR and ACC.

\subsubsection{Comparison on Poisoned Image Filtering}

First, we evaluate how many poisoned training images can be correctly filtered out with these methods. 

From the Table \ref{filter}, we can see that Spectral has the worst performance. And PBE  is better than SPECTRE. Nevertheless, it is clear that PUD significantly outperforms the other three methods.

Comparing the Table \ref{tb:dp} with the Table \ref{filter}, it is interesting to note that SPECTRE has a clear performance degradation as the number of poisoned images increases (from $500$ poisoned images to $5000$ poisoned images). In contrast, our prediction-based strategy (\emph{i.e.,} PBE in Table \ref{filter}) has a stable performance. Clearly, the SPECTRE outperforms PBE for the case of $500$ poisoned images, while the PBE outperforms the SPECTRE for the case of $5000$ poisoned images. It further indicates that the prediction-based strategy and SPECTRE are complementary.

\begin{table}[htpb]
\centering
\caption{\centering Number of filtered poisoned images on CIFAR-10. There are $5000$ poisoned images in total.} 
\label{filter}
 \resizebox{0.5\textwidth}{!}{
\begin{tabular}{ccccc}
    \toprule
     & Spectral & SPECTRE & PBE & PUD \\
    \midrule
    Badnet & 3465/5000 &4259/5000 & 4941/5000 & \textbf{4992}/5000  \\
    Blend  & 3168/5000 & 4824/5000 & 4960/5000 & \textbf{4996}/5000  \\
    SIG    & 3239/5000 & 4849/5000 &4956/5000 & \textbf{5000}/5000  \\
    \myblue{Blind}    & \myblue{3453/5000} & \myblue{4093/5000} & \myblue{4988/5000} & \myblue{\textbf{5000}/5000}  \\
    \myblue{AdvDoor}    & \myblue{3437/5000} & \myblue{2740/5000} & \myblue{4940/5000} & \myblue{\textbf{4988}/5000}  \\
    \bottomrule
\end{tabular}
}
\end{table}

\subsubsection{Comparison on Final Defense Performance}
The final defensive performance is also evaluated and compared. 
For Spectral Signatures and SPECTRE, we use their outputs (filtered training images) to train a new model. Comparisons are made between the new model from Spectral/SPECTRE and the purified model from PUD.

Table.\ref{df} and Table.\ref{df2} illustrate the comparisons of the defensive performance. Obviously, Spectral is too weak that it only has some defensive effect against WaNet attack. This is because WaNet emphasizes too much on stealth in the attack, so that it is easy to be defended against. SPECTRE can properly defend against Blend attack, but fail for Badnet and SIG attacks. 
In contrast, both PBE and PUD remarkably outperform Spectral and SPECTRE for all backdoor attacks. And there is a clear gain from PBE to PUD. It is due to the better data purification mechanism proposed by PUD.

\begin{table}[htpb]
\scriptsize
 \centering
\caption{\centering Final backdoor defense performance on CIFAR10.}
\label{defense}
 \resizebox{0.5\textwidth}{!}{
 \setlength{\tabcolsep}{0.3em}% 
\begin{tabular}{c|cc|cc|cc|cc|cc}
    \hline
        %&\multicolumn{2}{c|}{\textbf{Before}} \multicolumn{2}{c|}{\textbf{Spectral}} & \multicolumn{2}{c|}{\textbf{SPECTRE}} & \multicolumn{2}{c|}{\textbf{PBE}} & \multicolumn{2}{c|}{\textbf{PUD}}
          &\multicolumn{2}{c|}{\textbf{Before}} &  \multicolumn{2}{c|}{\textbf{Spectral}} & \multicolumn{2}{c|}{\textbf{SPECTRE}} & \multicolumn{2}{c|}{\textbf{PBE}} & \multicolumn{2}{c}{\textbf{PUD}} \\
    \cline{2-11}
     & ACC & ASR
    & ACC$\uparrow$ & ASR$\downarrow$
    & ACC$\uparrow$ & ASR$\downarrow$
    & ACC$\uparrow$ & ASR$\downarrow$  
    & ACC$\uparrow$ & ASR$\downarrow$ \\
    \hline
    Badnet  & 94.67 &98.97 & 89.61 & 98.07   &    93.43 &93.08         & 94.02 &11.30    &93.51 &\textbf{1.1}\\
    Blend   & 94.62 & 93.54 & 89.90 &99.98  & 93.39 & 39.67         &93.04 & 1.16   &94.26 &\textbf{0.91}  \\
    SIG     & 94.15&96.02 & 89.46 &99.81  & 92.89 & 98.17          &93.56 & 1.76 & 92.78 &\textbf{0.38}  \\
    WaNet   & 94.15 &99.50 & 91.35   &84.07   & 92.58    &23.89     &89.30&6.70   &94.03 &\textbf{0.99}\\
    \bottomrule
\end{tabular}
}
\label{df}
\end{table}

\begin{table}[htpb]
\scriptsize
 \centering
\caption{\centering Final backdoor defense performance on GTSRB.}
\label{defense}
 \resizebox{0.5\textwidth}{!}{
 \setlength{\tabcolsep}{0.3em}% 
\begin{tabular}{c|cc|cc|cc|cc|cc}
    \hline
        %&\multicolumn{2}{c|}{\textbf{Before}} \multicolumn{2}{c|}{\textbf{Spectral}} & \multicolumn{2}{c|}{\textbf{SPECTRE}} & \multicolumn{2}{c|}{\textbf{PBE}} & \multicolumn{2}{c|}{\textbf{PUD}}
          &\multicolumn{2}{c|}{\textbf{Before}} &  \multicolumn{2}{c|}{\textbf{Spectral}} & \multicolumn{2}{c|}{\textbf{SPECTRE}} & \multicolumn{2}{c|}{\textbf{PBE}} & \multicolumn{2}{c}{\textbf{PUD}} \\
    \cline{2-11}
     & ACC & ASR
    & ACC$\uparrow$ & ASR$\downarrow$
    & ACC$\uparrow$ & ASR$\downarrow$
    & ACC$\uparrow$ & ASR$\downarrow$  
    & ACC$\uparrow$ & ASR$\downarrow$ \\
    \hline
    Badnet  & 98.67 & 100.0 & 92.96 & 97.06   &    98.01 & 95.12         &90.13 & 12.31    &97.99 &\textbf{0.25} \\
    Blend   & 98.93&99.94 & 93.23&99.58    &    98.07 &45.31         &90.10&8.74   &97.43&\textbf{0.80}  \\
    SIG     & 97.92&99.81 & 94.70&99.91      &   97.82&56.42       &90.31&7.59  &98.29&\textbf{0.38} \\
    WaNet   & 98.97&98.78 & 96.76&92.32          &98.51& 43.08           & 91.23&3.63 &98.85&\textbf{0.24}\\
    \bottomrule
\end{tabular}
}
\label{df2}
\end{table}

\begin{table}[htpb]
\scriptsize
 \centering
\caption{\centering Final backdoor defense performance on sub-ImgNet1K.}
\label{defense}
 \resizebox{0.5\textwidth}{!}{
 \setlength{\tabcolsep}{0.3em}% 
\begin{tabular}{c|cc|cc|cc|cc|cc}
    \hline
        %&\multicolumn{2}{c|}{\textbf{Before}} \multicolumn{2}{c|}{\textbf{Spectral}} & \multicolumn{2}{c|}{\textbf{SPECTRE}} & \multicolumn{2}{c|}{\textbf{PBE}} & \multicolumn{2}{c|}{\textbf{PUD}}
          &\multicolumn{2}{c|}{\textbf{Before}} &  \multicolumn{2}{c|}{\textbf{Spectral}} & \multicolumn{2}{c|}{\textbf{SPECTRE}} & \multicolumn{2}{c|}{\textbf{PBE}} & \multicolumn{2}{c}{\textbf{PUD}} \\
    \cline{2-11}
     & ACC & ASR
    & ACC$\uparrow$ & ASR$\downarrow$
    & ACC$\uparrow$ & ASR$\downarrow$
    & ACC$\uparrow$ & ASR$\downarrow$  
    & ACC$\uparrow$ & ASR$\downarrow$ \\
    \hline
    Badnet  &97.37   &99.17  &98.11   &89.45     &    98.01 &91.46        &92.25  &4.01     &94.39  &\textbf{1.92 }\\
    Blend   &97.01  &97.90  &96.94  &88.36         &    97.73 & 16.98         & 94.50  &0.65     &95.08  &\textbf{0.29 }  \\
    SIG     &97.52  &99.91  & 96.12  & 98.64 & 97.45 &  91.15      & 94.39 & 0.62  & 95.79 &\textbf{0.30}  \\
    WaNet   &98.03   & 88.53  & 97.01  & 25.05 & 96.22   & 14.93  & 95.31 & 0.89    &96.43 &\textbf{0.46}\\
    \bottomrule
\end{tabular}
}
\label{df3}
\end{table}

\subsection{Robustness against Adversarial Attacks}\label{sec:adv}
Our approach can defend not only against backdoor attacks, but also against adversarial attacks. As we know, Adversarial Training (AT)~\cite{madry2018towards} is widely regarded as the most successful adversarial defense. Beside that, TRADES~\cite{TRADES} is another notable method, that can balance the trade-off between standard and robust accuracy. 

In this section, we take the PGD attack~\cite{madry2018towards} as a representative adversarial attack, and evaluate how much different defense methods can defend against the PGD attack. We compare our algorithm with two advanced adversarial defense methods, \emph{i.e.,} AT and TRADES in this paper.

\begin{table}[h]
\scriptsize
 \centering
\caption{\centering Robustness against adversarial attacks on CIFAR10.}
\label{defense}
 \resizebox{0.45\textwidth}{!}{
\begin{tabular}{ccccc}
    \toprule
         & Normal Training & AT & TRADES & PUD \\
    \midrule
    ACC  &94.01  &89.46  &   84.81 &92.32 \\ %\red{83.17}
    RACC   &0.01  &\textbf{77.49}  &   71.07 &76.13\\ %\red{89.21}
    \bottomrule
\end{tabular}
}
\label{AT}
\end{table}

\begin{table}[h]
\scriptsize
 \centering
\caption{\centering Robustness against adversarial attacks on GTSRB.}
\label{defense}
 \resizebox{0.45\textwidth}{!}{
\begin{tabular}{ccccc}
    \toprule
         & Normal Training & AT & TRADES & PUD \\
    \midrule
    ACC  &99.36  &92.16  &94.3    &98.45\\
    RACC   &11.80  &70.77  &\textbf{74.01}   &68.46 \\
    \bottomrule
\end{tabular}
}
\label{AT2}
\end{table}

\begin{table}[h]
\scriptsize
 \centering
\caption{\centering Robustness against adversarial attacks on sub-ImgNet1K.}
\label{defense}
 \resizebox{0.45\textwidth}{!}{
\begin{tabular}{ccccc}
    \toprule
         & Normal Training & AT & TRADES & PUD \\
    \midrule
    ACC  & 98.67 &94.31  & 91.26 &94.54   \\
    RACC   &3.64  &79.18  &77.75  &\textbf{79.89} \\
    \bottomrule
\end{tabular}
}
\label{AT3}
\end{table}

From Table \ref{AT}, Table \ref{AT2}, and Table \ref{AT3}, we can see that normal training cannot defend against adversarial attacks, \emph{i.e.,} poor performance in terms of RACC. For CIFAR-10 dataset, our approach is inferior to AT, but it is better than TRADES. For GTSRB dataset, our approach is inferior to TRADES, but it is competitive with AT. For sub-ImgNet1K dataset, our approach outperforms both AT and TRADES.
Therefore, it illustrates that our approach has a competitive ability to defend against adversarial attacks.

\section{Conclusion}\label{sec:conclusion}
In this work, we reveal that there is an intriguing connection between backdoor and adversarial attacks, \emph{i.e.}, for an infected model, its adversarial examples have similar features as its triggered images. We provide a theoretical analysis to explain such observations. Furthermore, inspired by such connections, we propose a unified defense method PUD which is able to defend against backdoor and adversarial attacks simultaneously. Even when an extra clean dataset is unavailable, our approach can still defend against state-of-the-art backdoor attacks (\emph{e.g.}, DynAtt, WaNet). 

Empirically, extensive experimental results show that our PUD outperforms both model repairing-based and data filtering-based backdoor defense methods. In terms of the robustness against adversarial attacks,
our approach has a competitive defense capability compared to adversarial training.
In the future, we hope that our discovery will encourage the community to jointly study backdoor and adversarial attacks.
%its adversarial examples are highly likely predicted as the backdoor target-label. Furthermore, we find that the features of adversarial images are very similar to that of triggered images. This tells us there is an underlying connections between adversarial and backdoor attacks, which inspires our community to jointly study them in the future. 

%------------------------------------------------------------------------
%------------------------------------------------------------------------
\section{Acknowledgements}
This work was supported partly by National Natural Science Foundations of China under grant No. 62272364, and Key Research and Development Program of Shaanxi (Program No.2024GX-YBXM-133 and No. 2024GH-ZDXM-47).
%-------------------------------------------------------------

%------------------------------------------------------------

% or


%\appendices

%\section{}

%\ifCLASSOPTIONcompsoc
%\else
%\fi

%This work was supported partly by National Key R\&D Program of China Grant 2018AAA0101400, NSFC Grants 61629301, 61773312, and 61976171, China Postdoctoral Science Foundation Grant 2019M653642, and Young Elite Scientists Sponsorship Program by CAST Grant 2018QNRC001.
\FloatBarrier
\bibliographystyle{abbrv}
\bibliography{main}
\end{document}